\documentclass[journal]{IEEEtran}

\usepackage{amsmath,amssymb}
\usepackage{graphicx}
\usepackage{algorithm}
\usepackage{algpseudocode}
\usepackage{booktabs}
\usepackage{cite}
\usepackage{multirow}
\usepackage{makecell}
\usepackage[hidelinks]{hyperref}
\usepackage{orcidlink}
\usepackage{xcolor}
\makeatletter
\g@addto@macro\normalsize{%
 \setlength\abovedisplayskip{3pt plus 2pt minus 1pt}%
 \setlength\belowdisplayskip{3pt plus 2pt minus 1pt}%
 \setlength\abovedisplayshortskip{1pt plus 1pt}%
 \setlength\belowdisplayshortskip{2pt plus 1pt minus 1pt}%
}
\makeatother

\usepackage{tikz}\usetikzlibrary{calc,shapes,arrows,automata,positioning,patterns}

\begin{document}

\title{\fontsize{20}{24}\selectfont
A Unified Mamba--MoE Surrogate for Closed-Loop Simulation\\ and
Measurement-Window Forecasting of Inverter Transients}

\author{Haoguang~Wang\,\orcidlink{0009-0005-4101-864X},~\IEEEmembership{}
 Huy~Hoang~Le\,\orcidlink{0009-0009-2138-5468},~\IEEEmembership{}
 Akhila~Kandivalasa\,\orcidlink{0009-0009-9066-6263},~\IEEEmembership{Student Member,~IEEE,}
 Christian~Moya\,\orcidlink{0000-0003-0180-9285},~\IEEEmembership{}
 Marcos~Netto\,\orcidlink{0000-0001-7002-3345},~\IEEEmembership{Senior Member,~IEEE,}
 and~Guang~Lin\textsuperscript{*}\,\orcidlink{0000-0002-0976-1987}~\IEEEmembership{}%
\thanks{
We acknowledge support from the National Science Foundation (DMS-2533878, DMS-2053746, DMS-2134209, ECCS-2328241, CBET-2347401, OAC-2311848); the Department of Energy Office of Science Advanced Scientific Computing Research program (DE-SC0023161) and Fusion Energy Sciences program (DE-SC0024583); and the SciDAC LEADS Institute.}
\thanks{H.~Wang and G.~Lin are with the School of Mechanical Engineering, Purdue University, West Lafayette, IN~47907~USA (e-mail: wang6411@purdue.edu; guanglin@purdue.edu). H. H. Le is with the R\&D Department, Powermore Ltd., Da Nang, 550000 VN (e-mail: lhhoang@powermore.vn). C.~Moya and G.~Lin are with the Department of Mathematics, Purdue University, West Lafayette, IN~47907~USA (e-mail: cmoyacal@purdue.edu). A.~Kandivalasa and M.~Netto are with the Department of Electrical and Computer Engineering, New Jersey Institute of Technology, Newark, NJ~07102~USA (e-mail: ak3366@njit.edu; marcos.netto@njit.edu). *Corresponding author: G.~Lin.}}

\markboth{}{}

\maketitle

\begin{abstract}
This paper proposes a Mamba surrogate model with mixture-of-experts (MoE) routing to represent the transient dynamics of inverter-based resources. A Mamba surrogate model is a predictive machine learning model built on the Mamba architecture. MoE routing uses a router network to assign data-dependent weights to specialized subnetworks (experts). The resulting Mamba--MoE surrogate can perform two tasks: (i) closed-loop simulation and (ii) measurement-window forecasting of inverter transients. A single Mamba backbone with task conditioning and expert routing serves both tasks, replacing two separate specialists. Task-matched objectives fit each prediction form, and an adaptive conformal layer provides prediction intervals for both tasks. For the considered grid-following inverter, the unified surrogate model remains in the same low-error regime as a Mamba specialist pair while using 13\% fewer parameters. The prediction intervals achieve 94--96\% empirical mean marginal coverage across the two tasks. For transient dynamics---that is, beyond the vicinity of an equilibrium point---our surrogate model with MoE routing yields lower errors across all outputs in both tasks compared to a shared Mamba backbone without expert routing. A controller hardware-in-the-loop simulation validates our results and shows that adapting only the shared output head with limited measured data reduces held-out forecasting error.
\end{abstract}


\IEEEpeerreviewmaketitle

\section{Introduction}
\label{sec:intro}

\IEEEPARstart{P}{ower} system dynamics are rapidly shifting toward being strongly driven by utility-scale inverter-based resources (IBRs). Indeed, IBR instantaneous penetration now regularly exceeds 60\% of total generation within the Electric Reliability Council of Texas's jurisdiction \cite{ERCOT}. This fundamental change has renewed interest in power system dynamics modeling \cite{Badrzadeh2024}, particularly regarding IBRs. The dynamics of IBRs, however, are difficult to represent with a first-principles model. 

IBRs' physical parameters vary with specific hardware and software implementations and with grid operating conditions. Different power-stage designs and control structures introduce nonidealities such as variable-frequency switching, control dead time, quantization and sampling errors, digital delay, and system parameter discrepancies. In addition, inverters are often equipped with nonlinear controllers to perform sophisticated functions, such as voltage ride-through, active power sharing, and frequency droop \cite{Li2022}. Therefore, first-principles modeling of inverters can be unreliable and inaccurate, especially when system parameters and control schemes are unknown or kept confidential \cite{Fan2022} due to strict nondisclosure requirements from original equipment manufacturers (OEMs) driven by security and intellectual property concerns.

Developing generic IBR models has also proven difficult. Parameterizing a generic IBR model is fundamentally a model-structure and identifiability problem \cite{Odunlami2025}: a reduced-order, nonproprietary model must reproduce dynamics governed largely by vendor-specific firmware, protection logic, current limits, and controller settings, which are often incompletely documented. The challenge is compounded by model flexibility---e.g., the Western Electricity Coordinating Council (WECC) central station photovoltaic power plant generic model \cite{WECC2015} has between 45–75 parameters and more than 30 possible control configurations. WECC explicitly cautions that the number of parameters available for tuning should be minimized to prevent mathematical degeneracy \cite{WECC2015}. Multiple parameter sets may fit a single disturbance yet be neither physically meaningful nor predictive under other conditions; parameterization requires selecting the correct control mode, estimating only identifiable parameters, and independently validating the fitted model \cite{WECC2026}.

Learning IBR models from terminal measurement data can capture input-output dynamics---including aggregation effects, control interactions, operating-point dependence, delays, and nonlinear limits---without requiring OEMs to disclose internal controller details. It can also provide computationally efficient representations for contingency screening, controller testing, real-time simulation, and operational decision support when detailed models are unavailable or too expensive to execute \cite{subedi2025survey}, especially when many scenarios must be screened repeatedly. This is the focus of the present work, and \S\ref{sec:related} provides a literature review. In particular, this paper asks whether a single learned surrogate model can serve two tasks: \emph{closed-loop simulation} and \emph{measurement-window forecasting}.

In closed-loop simulation, the goal is to perform an autoregressive rollout---that is, an inference method where a model's prediction for one step is fed back into itself as the input for the next step, repeating this loop iteratively to generate future time-series trajectories. In measurement-window forecasting, the goal is to use a fixed-size window of recent past data—the measurement or input window—to predict future values. By shifting, rolling, or sliding this window forward as new data arrives, the surrogate model should capture recent trends, adapt to structural shifts, and continuously output updated forecasts.

The tasks differ in their inputs and error propagation: \emph{simulation} advances the output trajectory step by step using control and reference inputs; hence, rollout errors can accumulate over the horizon \cite{venkatraman2015improving}, whereas \emph{forecasting} directly uses a recent measurement window to predict outputs over a horizon without auxiliary inputs. Because the same latent inverter dynamics generate both tasks and predict the same variables, albeit through different conditional prediction maps, training two dedicated models can duplicate the same dynamic information \cite{caruana1997multitask}. However, a plain shared network may not adapt well to both tasks. The central challenge is to exploit the common dynamic structure within a single model while preserving task-specific prediction behavior. This paper proposes a unified Mamba surrogate with mixture-of-experts (MoE) routing \cite{jacobs1991adaptive}, referred to as Mamba--MoE, for both tasks. 

The Mamba backbone is a selective state-space model (SSM) that recursively updates a hidden state to capture temporal dependencies in sampled transient trajectories \cite{gu2024mamba}. A task-conditioning mechanism indicates whether the current input is for \emph{simulation} or \emph{forecasting}, allowing the shared backbone to adapt its representation to the task identity \cite{perez2018film}. To reflect these two prediction forms, this paper uses task-matched training objectives and adds an adaptive conformal layer that equips each simulated and forecast trajectory with a calibrated prediction interval, enabling the surrogate to report its own uncertainty for downstream screening and risk assessment \cite{angelopoulos2021gentle, gibbs2021adaptive}. The MoE routing layer weights expert subnetworks by task type and operating condition, sharing dynamics while adapting across operating regimes \cite{shazeer2017outrageously}.

The contributions are threefold. First, this paper formulates inverter transient simulation and measurement-window forecasting as a dual-task surrogate modeling problem, making explicit their shared prediction target and distinct input structures. Second, it develops a unified model that combines a Mamba temporal backbone, task conditioning, and MoE routing, enabling a single network to serve both tasks rather than training two separate specialists. Third, it combines task-matched objectives with adaptive conformal prediction and evaluates error, parameter efficiency, interval coverage, robustness under operating-point shifts and measurement corruption, and cross-topology forecasting and adaptation. From a practical standpoint, this paper presents an early study with preliminary results on applying modern artificial intelligence (AI) to \emph{power system dynamics}. Realistic measurement data collected from a controller hardware-in-the-loop (CHIL) simulation are used to evaluate the forecasting task for cross-system adaptation.

This paper is organized as follows. Section \ref{sec:related} reviews related work. Section \ref{sec:system} describes the system model and defines the two tasks. Sections \ref{sec:arch}--\ref{sec:aci} present the model, training, and prediction-interval methods. Section \ref{sec:results} reports the experimental results, and Section \ref{sec:conclusion} concludes.

\section{Related Work}
\label{sec:related}

\subsection{Data-Driven Surrogates for IBR Dynamics}

Data-driven surrogates include recurrent, convolutional, operator-learning, and Gaussian-process models \cite{subedi2025survey}. Recent measurement-based studies also learn sub-cycle inverter-based resource dynamics from synchro-waveform measurements \cite{mohsenzadeh2025data}. Interface-aware long short-term memory models enforce terminal voltage-current consistency but require inverter equations that may be unavailable \cite{yang2025ibr}. Operator-learning methods, including DeepONet \cite{lu2021learning} and graph operator networks, have been used to predict post-fault swing equation trajectories \cite{moya2023deeponet, sun2023deepgraphonet}; DeepONet is therefore included as a baseline. These methods address single-input, single-output formulations or rely on physical constraints specific to a particular prediction setting. In contrast, this paper frames \emph{closed-loop simulation} and \emph{measurement-window forecasting} as distinct task formulations of the same inverter dynamics and asks whether a single surrogate serves both. Relative to established power-system tools, electromagnetic-transient models remain the accuracy reference \cite{Badrzadeh2024} (hence, we validate our surrogate using CHIL simulation); the surrogate instead targets repeated-screening regimes where detailed runs are computationally expensive or operational planning scenarios where calibrated models are unavailable. It is complementary to dynamic equivalencing \cite{chow2013coherency} and to forecasting-aided state estimation \cite{zhao2019dse}, with which the measurement-window forecasting task is most closely related. Physics-constrained surrogates are also strong contenders in data-limited regimes: Koopman-operator methods \cite{Susuki2016}, gray-box neural ordinary differential equations \cite{chen2018neural}, and low-order parametric or reduced-order models \cite{chow2013coherency} incorporate structure to improve data efficiency. The proposed physics-agnostic surrogate, by contrast, shares a single network across both tasks and is complementary to these approaches.

\subsection{Deep Learning Sequence Architectures for Time Series}

\emph{Structured} SSMs like S4~\cite{gu2022efficiently} use fixed, input-independent matrices for efficient parallel training via convolutions, while \emph{selective} SSMs (like Mamba~\cite{gu2024mamba}) make those parameters dynamically dependent on the input to filter information like an attention mechanism. In particular, PowerMamba~\cite{menati2024powermamba} applies Mamba for quasi-steady-state forecasting problems---specifically, load, electricity price, ancillary service price, and renewable generation forecasting---showing that selective SSMs are effective for power system time series. This paper focuses on \emph{power system dynamics}. In particular, IBR transient dynamics under different task formulations.

\subsection{Task Conditioning and Expert Routing}

To adapt a shared sequence backbone to different tasks, prior work commonly uses task conditioning or expert routing. Feature-wise linear modulation (FiLM) injects conditioning information through learned feature scaling and shifting~\cite{perez2018film}, while MoE layers condition the input through expert subnetworks~\cite{jacobs1991adaptive, shazeer2017outrageously}. Prior MoE-augmented Mamba models (MoE-Mamba~\cite{piero2024moeMamba}, BlackMamba~\cite{anthony2024blackmamba}, and Routing Mamba~\cite{routing2025mamba}) mainly increase capacity or training efficiency in large models via per-layer or sparse routing, rather than separating task-conditioned transient behavior in a compact power-system surrogate. We therefore use one compact dense soft-routing layer after the shared Mamba backbone, conditioned on task identity and trajectory features, so expert mixing serves the inverter simulation and forecasting tasks~\cite{puigcerver2023sparse}.

\section{System Model and Task Formulation}
\label{sec:system}

This section first presents an inverter--infinite-bus system (Fig. \ref{f1}) and its parameters, and then defines the closed-loop simulation and measurement-window forecasting tasks.

\subsection{Inverter--Infinite-Bus System}

\begin{figure}
\centering
\begin{tikzpicture}[line cap=round,line join=round,>=triangle 45,x=1cm,y=1cm,scale=4.0]
\clip(4.57,3.88) rectangle (6.6,4.11);
\draw [line width=0.4pt] (4.6,4)-- (4.7,4);
\draw [shift={(4.75,4)},line width=0.4pt]  plot[domain=0:3.141592653589793,variable=\t]({1*0.05*cos(\t r)+0*0.05*sin(\t r)},{0*0.05*cos(\t r)+1*0.05*sin(\t r)});
\draw [shift={(4.85,4)},line width=0.4pt]  plot[domain=0:3.141592653589793,variable=\t]({1*0.05*cos(\t r)+0*0.05*sin(\t r)},{0*0.05*cos(\t r)+1*0.05*sin(\t r)});
\draw [shift={(4.95,4)},line width=0.4pt]  plot[domain=0:3.141592653589793,variable=\t]({1*0.05*cos(\t r)+0*0.05*sin(\t r)},{0*0.05*cos(\t r)+1*0.05*sin(\t r)});
\draw [line width=0.4pt] (5.1,4)-- (5.12,4.04);
\draw [line width=0.4pt] (5.12,4.04)-- (5.16,3.96);
\draw [line width=0.4pt] (5.16,3.96)-- (5.2,4.04);
\draw [line width=0.4pt] (5.2,4.04)-- (5.24,3.96);
\draw [line width=0.4pt] (5.24,3.96)-- (5.28,4.04);
\draw [line width=0.4pt] (5,4)-- (5.1,4);
\draw [shift={(5.55,4)},line width=0.4pt]  plot[domain=0:3.141592653589793,variable=\t]({1*0.05*cos(\t r)+0*0.05*sin(\t r)},{0*0.05*cos(\t r)+1*0.05*sin(\t r)});
\draw [shift={(5.65,4)},line width=0.4pt]  plot[domain=0:3.141592653589793,variable=\t]({1*0.05*cos(\t r)+0*0.05*sin(\t r)},{0*0.05*cos(\t r)+1*0.05*sin(\t r)});
\draw [shift={(5.75,4)},line width=0.4pt]  plot[domain=0:3.141592653589793,variable=\t]({1*0.05*cos(\t r)+0*0.05*sin(\t r)},{0*0.05*cos(\t r)+1*0.05*sin(\t r)});
\draw [line width=0.4pt] (5.8,4)-- (5.9,4);
\draw [line width=0.4pt] (5.9,4)-- (5.92,4.04);
\draw [line width=0.4pt] (5.92,4.04)-- (5.96,3.96);
\draw [line width=0.4pt] (5.96,3.96)-- (6,4.04);
\draw [line width=0.4pt] (6,4.04)-- (6.04,3.96);
\draw [line width=0.4pt] (6.04,3.96)-- (6.08,4.04);
\draw [line width=0.4pt] (6.3,4.1)-- (6.3,3.9);
\draw [line width=0.4pt] (6.3,3.9)-- (6.5,3.9);
\draw [line width=0.4pt] (6.5,3.9)-- (6.5,4.1);
\draw [line width=0.4pt] (6.5,4.1)-- (6.3,4.1);
\draw [line width=0.4pt] (6.3,3.9)-- (6.5,4.1);
\draw [line width=0.4pt] (6.3,4)-- (6.18,4);
\draw [line width=0.4pt] (6.227118174711503,4) -- (6.24,4.017175767051434);
\draw [line width=0.4pt] (6.227118174711503,4) -- (6.24,3.9828242329485657);
\draw [line width=0.4pt] (6.18,4)-- (6.1,4);
\draw [line width=0.4pt] (6.1,4)-- (6.08,4.04);
\draw [line width=0.4pt] (5.5,4)-- (5.36,4);
\draw [line width=0.4pt] (5.417118174711503,4) -- (5.43,4.017175767051434);
\draw [line width=0.4pt] (5.417118174711503,4) -- (5.43,3.9828242329485657);
\draw [line width=0.4pt] (5.36,4)-- (5.3,4);
\draw [line width=0.4pt] (5.28,4.04)-- (5.3,4);
\begin{scriptsize}
\draw [fill=black] (4.6,4) circle (0.4pt);
\draw [fill=black] (6.16,4) circle (0.4pt);
\draw [fill=black] (5.36,4) circle (0.4pt);
\end{scriptsize}
\draw (4.76,4.00) node[anchor=north west] {$r_{g}+j\omega\ell_{g}$};
\draw (5.55,4.00) node[anchor=north west] {$r_{f}+j\omega\ell_{f}$};
\draw (6.15,4.00) node[anchor=north west] {\footnotesize{$\bar{i}^{\mathrm{cv}}$}};
\draw (6.11,4.12) node[anchor=north west] {\footnotesize{$\bar{v}^{\mathrm{cv}}$}};
\draw (5.34,4.00) node[anchor=north west] {\footnotesize{$\bar{i}^{\mathrm{filt}}$}};
\draw (5.30,4.114) node[anchor=north west] {\footnotesize{$\bar{v}_{1}$}};
\draw (4.54,4.12) node[anchor=north west] {\footnotesize{$\bar{v}^{\mathrm{grid}}$}};
\draw (6.285,4.105) node[anchor=north west] {\footnotesize{AC}};
\draw (6.355,4.000) node[anchor=north west] {\footnotesize{DC}};
\end{tikzpicture}
\vspace{-.2cm}
\caption{One-line diagram of the inverter--infinite-bus system. The inverter connects to the grid (modeled as an infinite bus) through a series $r\ell$ branch, with $\bar{v}^{\mathrm{cv}}=v^{\mathrm{cv}}e^{j\theta^{\mathrm{cv}}}$ and $\bar{i}^{\mathrm{cv}}=i^{\mathrm{cv}}e^{j\alpha^{\mathrm{cv}}}$.}
\label{f1}
\end{figure}
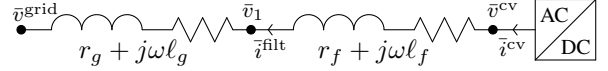

The inverter in Fig. \ref{f1} is a standard grid-following inverter \cite{Li2022} equipped with a synchronization controller---that is, a phase-locked loop (PLL); an outer (power) controller; an inner (current) controller; and a dc-link voltage controller. We represent the system using an averaged model; see \cite{Zhao2026} for the explicit nonlinear differential--algebraic equations. The branch current is modeled as an algebraic variable rather than as an independent differential state. 

The system model is expressed in per unit (pu) on the inverter base power $S_{\mathrm{gen}}=2.75$\,MVA at 60\,Hz. The system base $S_{\mathrm{sys}}=100$\,MVA and the inverter base use the same nominal base voltage; hence, the $0.075$-pu grid reactance on the system base corresponds to $\omega\ell_g = 0.075(S_{\mathrm{gen}}/S_{\mathrm{sys}}) = 0.00206$\,pu on the inverter base. The grid resistance $r_g=0$, and the Bus~1 voltage magnitude $|v_1|=1.00001$\,pu is constant since the grid is modeled as an infinite bus; together, these settings define a single-inverter strong-grid condition. The filter impedance $z_f=r_f+j\omega\ell_f=0.016+j0.009$\,pu. The dc-link voltage is regulated at $V_{\mathrm{dc}}=1.0$\,pu; PLL-, power-, and current-loop (proportional, integral) controller gains are $(2.0,20.0)$, $(2.0,30.0)$, and $(0.37,0.7)$, respectively. The PLL input low-pass filter uses $\omega_{\mathrm{\ell p}}=2\pi\times66$\,rad/s, the active- and reactive-power measurement filters use $\omega_z=\omega_f=2\pi\times6.6$\,rad/s, and the voltage-feedforward gain $k_{\mathrm{ffv}}=0$. Note that although we develop the Mamba--MoE surrogate using an inverter--infinite-bus system, we report results for a numerically simulated 9-bus system and for data collected from a realistic CHIL simulation \cite{Orsinger2021}; see \S\ref{sec:results} for more details. We select the following four current components to form the observed output vector:
\begin{equation}
x_t = \bigl[i^{\mathrm{cv}}_d,\; i^{\mathrm{cv}}_q,\; i_{r}^{\mathrm{filt}},\; i_{i}^{\mathrm{filt}}\bigr]^\mathsf{T}
\in \mathbb{R}^{n_x},\qquad n_x=4,
\label{eq:state}
\end{equation}
where $r$ and $i$ denote real and imaginary components in the network reference frame, whereas $d$ and $q$ denote direct and quadrature components aligned with the PLL $dq$ reference frame. These outputs are two coordinate representations of the same branch current. For a nonzero branch current, the PLL angle $\theta_{\mathrm{PLL}}$ is not included as a separate output; it equals the network-frame current phase minus the $dq$-frame current phase, modulo $2\pi$. The controller-integrator variables are not included in $x_t$.

\subsection{Unified Task Formulation}
\label{sec:modes}

Accordingly, $x_t$ is an observed output vector rather than a complete Markov state; the task mappings below are conditioned on encoded observation history and recurrent state. Both tasks use the same $x_t$ but differ in their inputs and prediction forms: simulation generates a trajectory by step-by-step rollout, whereas forecasting produces $H$ future outputs from a recent measurement window in a single pass. Hereafter, the superscripts $\mathrm{sim}$ and $\mathrm{fcast}$ are used only in mathematical notation to denote the simulation and forecasting tasks, respectively; hats denote model predictions.

\textbf{Simulation task.} At each autoregressive step $t$, the model receives the current-output vector, the active-power output, $p_m\in\mathbb{R}$, and the time-varying active-power reference $p_\mathrm{ref}(t)$:
\begin{equation}
\xi^\mathrm{sim}_t = [x_t,\, p_m,\, p_\mathrm{ref}(t)] \in \mathbb{R}^{n_x+2}.
\label{eq:sim_input}
\end{equation}
The one-step residual target is defined as $\delta x_t=x_{t+1}-x_t$ and obtained directly from the sampled data rather than derived from the continuous-time equations. Encoding a known history window initializes a latent state that evolves across rollout steps. The rollout begins from the window's last observed output, and each predicted output is reused as the next input. The model predicts $r_{1,t}\in\mathbb{R}^{n_x}$ as an estimate of $\delta x_t$ and advances the output as
\begin{equation}
\hat{x}^\mathrm{sim}_{t+1}
= \hat{x}^\mathrm{sim}_t + r_{1,t}.
\label{eq:sim_pred}
\end{equation}
Because the predicted output is fed back as a subsequent input, simulation error may accumulate with the rollout horizon.

\textbf{Forecasting task.} This task targets short-term current-output forecasting from local measurements when the control references of black-box IBRs are unavailable to the system operator. It uses a window of the most recent $W$ current-output samples as input. The power command and reference input used in the simulation task are not provided to the forecasting task; to keep a unified input dimension, the two auxiliary channels are set to zero at each time $\tau$ in the window:
\begin{equation}
\xi^\mathrm{fcast}_\tau = [x_\tau,\, 0,\,0]
\in \mathbb{R}^{n_x+2},\quad \tau=t-W+1,\ldots,t.
\label{eq:forec_input}
\end{equation}
These zero-filled auxiliary channels serve as a constant null/missingness marker rather than a physical command value ($p_m$ lies in $[0.4,0.8]$\,pu, so zero is non-physical), and the forecasting mode is jointly identified by the task embedding $c_m$ introduced in \S\ref{sec:arch}. The model produces $H$ future residuals $[r_1,\ldots,r_H] \in \mathbb{R}^{H\times n_x}$ in a single forward pass. Each residual $r_h$ approximates the displacement from the last observed output $x_t$ to the future output $x_{t+h}$:
\begin{equation}
r_h \approx x_{t+h} - x_t,\qquad
\hat{x}^\mathrm{fcast}_{t+h} = x_t + r_h,\quad h=1,\ldots,H.
\label{eq:forec_pred}
\end{equation}
This direct $H$-output prediction form does not recursively use predicted outputs as inputs and is therefore distinct from autoregressive simulation rollout. Because the forecaster receives no future commands, two cases that share the same past window but differ in future references are indistinguishable; forecasting is therefore evaluated under the fixed reference-step timing used to generate the offline data, rather than under arbitrary unobserved command changes. Fig.~\ref{f2} illustrates the simulation and forecasting tasks.

\section{Proposed Mamba--MoE Surrogate Framework}
\label{sec:arch}

\subsection{Architecture Overview}\label{arch_desc}

\begin{figure}[!t]
\centering
\definecolor{wwwwww}{rgb}{0.4,0.4,0.4}
\definecolor{zzttqq}{rgb}{0.6,0.2,0}
\definecolor{qqwuqq}{rgb}{0,0.39215686274509803,0}
\begin{tikzpicture}[line cap=round,line join=round,>=triangle 45,x=1cm,y=1cm]
\clip(-0.01,-0.01) rectangle (8.51,3.51);
\fill[line width=0.4pt,color=qqwuqq,fill=qqwuqq,fill opacity=0.1] (0,0) -- (1.5,0) -- (1.5,1) -- (0,1) -- cycle;
\fill[line width=0.4pt,color=qqwuqq,fill=qqwuqq,fill opacity=0.1] (0.6,1.1) -- (1.9,1.1) -- (1.9,1.7) -- (0.6,1.7) -- cycle;
\fill[line width=0.4pt,color=qqwuqq,fill=qqwuqq,fill opacity=0.1] (0,2.5) -- (1.7,2.5) -- (1.7,3.5) -- (0,3.5) -- cycle;
\fill[line width=0.4pt,color=qqwuqq,fill=qqwuqq,fill opacity=0.1] (0.6,1.8) -- (1.8,1.8) -- (1.8,2.4) -- (0.6,2.4) -- cycle;
\fill[line width=0.4pt,color=qqwuqq,fill=qqwuqq,fill opacity=0.1] (2.1,1.7) -- (3.3,1.7) -- (3.3,3.5) -- (2.1,3.5) -- cycle;
\fill[line width=0.4pt,color=qqwuqq,fill=qqwuqq,fill opacity=0.1] (2.1,0) -- (3.3,0) -- (3.3,0.4) -- (2.1,0.4) -- cycle;
\fill[line width=0.4pt,color=zzttqq,fill=zzttqq,fill opacity=0.1] (3.6,1.5) -- (4.8,1.5) -- (4.8,3.5) -- (3.6,3.5) -- cycle;
\fill[line width=0.4pt,color=zzttqq,fill=zzttqq,fill opacity=0.1] (3.7,1.6) -- (4.7,1.6) -- (4.7,1.9) -- (3.7,1.9) -- cycle;
\fill[line width=0.4pt,color=zzttqq,fill=zzttqq,fill opacity=0.1] (3.7,2.4) -- (4.7,2.4) -- (4.7,2.7) -- (3.7,2.7) -- cycle;
\fill[line width=0.4pt,color=zzttqq,fill=zzttqq,fill opacity=0.1] (3.7,2.8) -- (4.7,2.8) -- (4.7,3.1) -- (3.7,3.1) -- cycle;
\fill[line width=0.4pt,color=qqwuqq,fill=qqwuqq,fill opacity=0.1] (2.6,0.6) -- (3.6,0.6) -- (3.6,1.2) -- (2.6,1.2) -- cycle;
\fill[line width=0.4pt,color=wwwwww,fill=wwwwww,fill opacity=0.1] (5,2.5) -- (6.5,2.5) -- (6.5,3.5) -- (5,3.5) -- cycle;
\fill[line width=0.4pt,color=wwwwww,fill=wwwwww,fill opacity=0.1] (6.7,2.5) -- (8.2,2.5) -- (8.2,3.5) -- (6.7,3.5) -- cycle;
\fill[line width=0.4pt,color=qqwuqq,fill=qqwuqq,fill opacity=0.1] (5.2,0.9) -- (6.4,0.9) -- (6.4,1.9) -- (5.2,1.9) -- cycle;
\fill[line width=0.4pt,color=qqwuqq,fill=qqwuqq,fill opacity=0.1] (7,1.2) -- (8,1.2) -- (8,2.1) -- (7,2.1) -- cycle;
\fill[line width=0.4pt,color=qqwuqq,fill=qqwuqq,fill opacity=0.1] (5.2,0) -- (6.4,0) -- (6.4,0.6) -- (5.2,0.6) -- cycle;
\fill[line width=0.4pt,color=qqwuqq,fill=qqwuqq,fill opacity=0.1] (6.8,0) -- (8,0) -- (8,1) -- (6.8,1) -- cycle;
\draw [line width=0.4pt,color=qqwuqq] (0,0)-- (1.5,0);
\draw [line width=0.4pt,color=qqwuqq] (1.5,0)-- (1.5,1);
\draw [line width=0.4pt,color=qqwuqq] (1.5,1)-- (0,1);
\draw [line width=0.4pt,color=qqwuqq] (0,1)-- (0,0);
\draw [line width=0.4pt,color=qqwuqq] (0.6,1.1)-- (1.9,1.1);
\draw [line width=0.4pt,color=qqwuqq] (1.9,1.1)-- (1.9,1.7);
\draw [line width=0.4pt,color=qqwuqq] (1.9,1.7)-- (0.6,1.7);
\draw [line width=0.4pt,color=qqwuqq] (0.6,1.7)-- (0.6,1.1);
\draw [line width=0.4pt,color=qqwuqq] (0,2.5)-- (1.7,2.5);
\draw [line width=0.4pt,color=qqwuqq] (1.7,2.5)-- (1.7,3.5);
\draw [line width=0.4pt,color=qqwuqq] (1.7,3.5)-- (0,3.5);
\draw [line width=0.4pt,color=qqwuqq] (0,3.5)-- (0,2.5);
\draw [line width=0.4pt,color=qqwuqq] (0.6,1.8)-- (1.8,1.8);
\draw [line width=0.4pt,color=qqwuqq] (1.8,1.8)-- (1.8,2.4);
\draw [line width=0.4pt,color=qqwuqq] (1.8,2.4)-- (0.6,2.4);
\draw [line width=0.4pt,color=qqwuqq] (0.6,2.4)-- (0.6,1.8);
\draw [line width=0.4pt] (0.2,2.5)-- (0.2,2.2);
\draw [line width=0.4pt] (0.2,2.2)-- (0.6,2.2);
\draw [line width=0.4pt] (0.2,1)-- (0.2,1.4);
\draw [line width=0.4pt] (0.2,1.4)-- (0.6,1.4);
\draw [line width=0.4pt,color=qqwuqq] (2.1,1.7)-- (3.3,1.7);
\draw [line width=0.4pt,color=qqwuqq] (3.3,1.7)-- (3.3,3.5);
\draw [line width=0.4pt,color=qqwuqq] (3.3,3.5)-- (2.1,3.5);
\draw [line width=0.4pt,color=qqwuqq] (2.1,3.5)-- (2.1,1.7);
\draw [line width=0.4pt,dash pattern=on 1pt off 1pt] (1.9,1.4)-- (2.9,1.4);
\draw [line width=0.4pt] (1.8,2.2)-- (2.1,2.2);
\draw [line width=0.4pt,dash pattern=on 1pt off 1pt] (2.9,1.4)-- (2.9,1.7);
\draw [line width=0.4pt,color=qqwuqq] (2.1,0)-- (3.3,0);
\draw [line width=0.4pt,color=qqwuqq] (3.3,0)-- (3.3,0.4);
\draw [line width=0.4pt,color=qqwuqq] (3.3,0.4)-- (2.1,0.4);
\draw [line width=0.4pt,color=qqwuqq] (2.1,0.4)-- (2.1,0);
\draw [line width=0.4pt] (1.7,1.1)-- (1.7,0.2);
\draw [line width=0.4pt] (1.7,0.2)-- (2.1,0.2);
\draw [line width=0.4pt,color=zzttqq] (3.6,1.5)-- (4.8,1.5);
\draw [line width=0.4pt,color=zzttqq] (4.8,1.5)-- (4.8,3.5);
\draw [line width=0.4pt,color=zzttqq] (4.8,3.5)-- (3.6,3.5);
\draw [line width=0.4pt,color=zzttqq] (3.6,3.5)-- (3.6,1.5);
\draw [line width=0.4pt,color=zzttqq] (3.7,1.6)-- (4.7,1.6);
\draw [line width=0.4pt,color=zzttqq] (4.7,1.6)-- (4.7,1.9);
\draw [line width=0.4pt,color=zzttqq] (4.7,1.9)-- (3.7,1.9);
\draw [line width=0.4pt,color=zzttqq] (3.7,1.9)-- (3.7,1.6);
\draw [line width=0.4pt,color=zzttqq] (3.7,2.4)-- (4.7,2.4);
\draw [line width=0.4pt,color=zzttqq] (4.7,2.4)-- (4.7,2.7);
\draw [line width=0.4pt,color=zzttqq] (4.7,2.7)-- (3.7,2.7);
\draw [line width=0.4pt,color=zzttqq] (3.7,2.7)-- (3.7,2.4);
\draw [line width=0.4pt,color=zzttqq] (3.7,2.8)-- (4.7,2.8);
\draw [line width=0.4pt,color=zzttqq] (4.7,2.8)-- (4.7,3.1);
\draw [line width=0.4pt,color=zzttqq] (4.7,3.1)-- (3.7,3.1);
\draw [line width=0.4pt,color=zzttqq] (3.7,3.1)-- (3.7,2.8);
\draw [line width=0.4pt,color=qqwuqq] (2.6,0.6)-- (3.6,0.6);
\draw [line width=0.4pt,color=qqwuqq] (3.6,0.6)-- (3.6,1.2);
\draw [line width=0.4pt,color=qqwuqq] (3.6,1.2)-- (2.6,1.2);
\draw [line width=0.4pt,color=qqwuqq] (2.6,1.2)-- (2.6,0.6);
\draw [line width=0.4pt] (3.1,1.7)-- (3.1,1.2);
\draw [line width=0.4pt] (2.9,0.6)-- (2.9,0.4);
\draw [line width=0.4pt] (3.6,0.9)-- (3.9,0.9);
\draw [line width=0.4pt] (3.9,0.9)-- (3.9,1.5);
\draw [line width=0.4pt,color=wwwwww] (5,2.5)-- (6.5,2.5);
\draw [line width=0.4pt,color=wwwwww] (6.5,2.5)-- (6.5,3.5);
\draw [line width=0.4pt,color=wwwwww] (6.5,3.5)-- (5,3.5);
\draw [line width=0.4pt,color=wwwwww] (5,3.5)-- (5,2.5);
\draw [line width=0.4pt,color=wwwwww] (6.7,2.5)-- (8.2,2.5);
\draw [line width=0.4pt,color=wwwwww] (8.2,2.5)-- (8.2,3.5);
\draw [line width=0.4pt,color=wwwwww] (8.2,3.5)-- (6.7,3.5);
\draw [line width=0.4pt,color=wwwwww] (6.7,3.5)-- (6.7,2.5);
\draw [line width=0.4pt,color=qqwuqq] (5.2,0.9)-- (6.4,0.9);
\draw [line width=0.4pt,color=qqwuqq] (6.4,0.9)-- (6.4,1.9);
\draw [line width=0.4pt,color=qqwuqq] (6.4,1.9)-- (5.2,1.9);
\draw [line width=0.4pt,color=qqwuqq] (5.2,1.9)-- (5.2,0.9);
\draw [line width=0.4pt,color=qqwuqq] (7,1.2)-- (8,1.2);
\draw [line width=0.4pt,color=qqwuqq] (8,1.2)-- (8,2.1);
\draw [line width=0.4pt,color=qqwuqq] (8,2.1)-- (7,2.1);
\draw [line width=0.4pt,color=qqwuqq] (7,2.1)-- (7,1.2);
\draw [line width=0.4pt,color=qqwuqq] (5.2,0)-- (6.4,0);
\draw [line width=0.4pt,color=qqwuqq] (6.4,0)-- (6.4,0.6);
\draw [line width=0.4pt,color=qqwuqq] (6.4,0.6)-- (5.2,0.6);
\draw [line width=0.4pt,color=qqwuqq] (5.2,0.6)-- (5.2,0);
\draw [line width=0.4pt,color=qqwuqq] (6.8,0)-- (8,0);
\draw [line width=0.4pt,color=qqwuqq] (8,0)-- (8,1);
\draw [line width=0.4pt,color=qqwuqq] (8,1)-- (6.8,1);
\draw [line width=0.4pt,color=qqwuqq] (6.8,1)-- (6.8,0);
\draw [line width=0.4pt,dash pattern=on 1pt off 1pt] (3.3,0.2)-- (5.2,0.2);
\draw [line width=0.4pt] (4.6,1.5)-- (4.6,0.4);
\draw [line width=0.4pt] (4.6,0.4)-- (5.2,0.4);
\draw [line width=0.4pt] (6.4,0.3)-- (6.8,0.3);
\draw [line width=0.4pt] (7.5,1)-- (7.5,1.2);
\draw [line width=0.4pt] (7,1.6)-- (6.4,1.6);
\draw [line width=0.4pt] (5.6,1.9)-- (5.6,2.5);
\draw [line width=0.4pt] (6,1.9)-- (6,2.3);
\draw [line width=0.4pt] (6,2.3)-- (7.4,2.3);
\draw [line width=0.4pt] (7.4,2.3)-- (7.4,2.5);
\begin{scriptsize}
\draw [fill=black,shift={(0.6,2.2)},rotate=270] (0,0) ++(0 pt,1.5pt) -- ++(1.299038105676658pt,-2.25pt)--++(-2.598076211353316pt,0 pt) -- ++(1.299038105676658pt,2.25pt);
\draw [fill=black,shift={(0.6,1.4)},rotate=270] (0,0) ++(0 pt,1.5pt) -- ++(1.299038105676658pt,-2.25pt)--++(-2.598076211353316pt,0 pt) -- ++(1.299038105676658pt,2.25pt);
\draw [fill=black,shift={(2.1,2.2)},rotate=270] (0,0) ++(0 pt,1.5pt) -- ++(1.299038105676658pt,-2.25pt)--++(-2.598076211353316pt,0 pt) -- ++(1.299038105676658pt,2.25pt);
\draw [fill=black,shift={(2.9,1.7)}] (0,0) ++(0 pt,1.5pt) -- ++(1.299038105676658pt,-2.25pt)--++(-2.598076211353316pt,0 pt) -- ++(1.299038105676658pt,2.25pt);
\draw [fill=black,shift={(2.1,0.2)},rotate=270] (0,0) ++(0 pt,1.5pt) -- ++(1.299038105676658pt,-2.25pt)--++(-2.598076211353316pt,0 pt) -- ++(1.299038105676658pt,2.25pt);
\draw [fill=black,shift={(3.1,1.2)},rotate=180] (0,0) ++(0 pt,1.5pt) -- ++(1.299038105676658pt,-2.25pt)--++(-2.598076211353316pt,0 pt) -- ++(1.299038105676658pt,2.25pt);
\draw [fill=black,shift={(2.9,0.4)},rotate=180] (0,0) ++(0 pt,1.5pt) -- ++(1.299038105676658pt,-2.25pt)--++(-2.598076211353316pt,0 pt) -- ++(1.299038105676658pt,2.25pt);
\draw [fill=black,shift={(3.9,1.5)}] (0,0) ++(0 pt,1.5pt) -- ++(1.299038105676658pt,-2.25pt)--++(-2.598076211353316pt,0 pt) -- ++(1.299038105676658pt,2.25pt);
\draw [fill=black,shift={(5.2,0.2)},rotate=270] (0,0) ++(0 pt,1.5pt) -- ++(1.299038105676658pt,-2.25pt)--++(-2.598076211353316pt,0 pt) -- ++(1.299038105676658pt,2.25pt);
\draw [fill=black,shift={(5.2,0.4)},rotate=270] (0,0) ++(0 pt,1.5pt) -- ++(1.299038105676658pt,-2.25pt)--++(-2.598076211353316pt,0 pt) -- ++(1.299038105676658pt,2.25pt);
\draw [fill=black,shift={(6.8,0.3)},rotate=270] (0,0) ++(0 pt,1.5pt) -- ++(1.299038105676658pt,-2.25pt)--++(-2.598076211353316pt,0 pt) -- ++(1.299038105676658pt,2.25pt);
\draw [fill=black,shift={(7.5,1.2)}] (0,0) ++(0 pt,1.5pt) -- ++(1.299038105676658pt,-2.25pt)--++(-2.598076211353316pt,0 pt) -- ++(1.299038105676658pt,2.25pt);
\draw [fill=black,shift={(6.4,1.6)},rotate=90] (0,0) ++(0 pt,1.5pt) -- ++(1.299038105676658pt,-2.25pt)--++(-2.598076211353316pt,0 pt) -- ++(1.299038105676658pt,2.25pt);
\draw [fill=black,shift={(5.6,2.5)}] (0,0) ++(0 pt,1.5pt) -- ++(1.299038105676658pt,-2.25pt)--++(-2.598076211353316pt,0 pt) -- ++(1.299038105676658pt,2.25pt);
\draw [fill=black,shift={(7.4,2.5)}] (0,0) ++(0 pt,1.5pt) -- ++(1.299038105676658pt,-2.25pt)--++(-2.598076211353316pt,0 pt) -- ++(1.299038105676658pt,2.25pt);
\end{scriptsize}
\draw (-0.03,0.98) node[anchor=north west] {\parbox{2.0 cm}{\scriptsize Task label:\\ \scriptsize simulation or\\ forecasting}};
\draw (-0.03,3.48) node[anchor=north west] {\parbox{2.0 cm}{\scriptsize Input sequence:\\ \scriptsize outputs and\\ auxiliary inputs}};
\draw (0.60,1.75) node[anchor=north west] {\parbox{2.0 cm}{\scriptsize Task\\ \scriptsize embedding}};
\draw (1.83,1.75) node[anchor=north west] {\parbox{2.0 cm}{\tiny conditioning}};
\draw (2.20,0.40) node[anchor=north west] {\parbox{2.0 cm}{\scriptsize Router}};
\draw (3.30,0.55) node[anchor=north west] {\parbox{2.0 cm}{\tiny expert weights}};
\draw (5.20,0.65) node[anchor=north west] {\parbox{2.0 cm}{\scriptsize Weighted \\ sum}};
\draw (5.25,1.75) node[anchor=north west] {\parbox{2.0 cm}{\scriptsize Output \\ residuals}};
\draw (5.10,3.55) node[anchor=north west] {\parbox{2.0 cm}{\scriptsize Simulation \\ output}};
\draw (5.00,2.95) node[anchor=north west] {\parbox{2.0 cm}{\tiny Next-step residual}};
\draw (6.75,2.95) node[anchor=north west] {\parbox{2.0 cm}{\tiny $H$-step residuals}};
\draw (6.75,3.55) node[anchor=north west] {\parbox{2.0 cm}{\scriptsize Forecasting \\ output}};
\draw (6.90,1.00) node[anchor=north west] {\parbox{2.0 cm}{\scriptsize Output \\ Mamba \\ layer}};
\draw (7.05,2.15) node[anchor=north west] {\parbox{2.0 cm}{\scriptsize Linear \\ output \\ head}};
\draw (2.60,1.25) node[anchor=north west] {\parbox{2.0 cm}{\scriptsize Routed \\ \scriptsize features}};
\draw (0.6,2.45) node[anchor=north west] {\parbox{2.0 cm}{\scriptsize Input\\ \scriptsize projection}};
\draw (2.10,3.10) node[anchor=north west] {\parbox{2.0 cm}{\scriptsize \textbf{Mamba}\\ \scriptsize backbone}};
\draw (2.25,2.25) node[anchor=north west] {\parbox{2.0 cm}{\tiny FiLM--\\ \tiny conditioned}};
\draw (3.52,3.52) node[anchor=north west] {\parbox{2.0 cm}{\scriptsize \textbf{K Experts}}};
\draw (3.62,3.18) node[anchor=north west] {\parbox{2.0 cm}{\scriptsize Expert 1}};
\draw (3.62,2.78) node[anchor=north west] {\parbox{2.0 cm}{\scriptsize Expert 2}};
\draw (3.62,1.98) node[anchor=north west] {\parbox{2.0 cm}{\scriptsize Expert K}};
\draw (4.02,2.64) node[anchor=north west] {\parbox{2.0 cm}{\scriptsize $\boldsymbol{\vdots}$}};
\end{tikzpicture}
\vspace{-.2cm}
\caption{The proposed Mamba--MoE surrogate architecture.}
\label{fig:arch}
\end{figure}
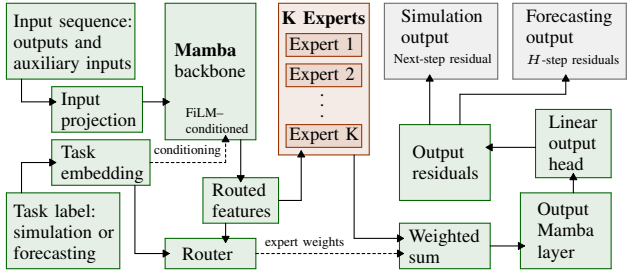

Fig.~\ref{fig:arch} shows the unified Mamba--MoE forward path. The input sequence is first projected into a $d_{\mathrm{model}}$-dimensional representation space. A binary task label $m\in\{0,1\}$ ($m{=}0$ simulation, $m{=}1$ forecasting) is mapped to a task embedding $c_m\in\mathbb{R}^{d_c}$. This embedding is used to apply FiLM to the Mamba blocks and, together with the routing representation, is fed to the MoE router. The Mamba backbone produces time-step hidden representations $h_t$; the expert-mixed output is then passed to a shared output Mamba block and a unified output head to generate $H$-step output residuals.

\begin{figure}[!t]
\centering
\definecolor{qqqqff}{rgb}{0,0,1}
\definecolor{cxvqqq}{rgb}{0.7803921568627451,0.3137254901960784,0}
\definecolor{wwwwww}{rgb}{0.4,0.4,0.4}
\definecolor{qqwuqq}{rgb}{0,0.39215686274509803,0}
\begin{tikzpicture}[line cap=round,line join=round,>=triangle 45,x=1cm,y=1cm,scale=4]
\clip(-.17,-.45) rectangle (2.1,0.36);
\fill[line width=0.4pt,color=qqwuqq,fill=qqwuqq,fill opacity=0.1] (0.24,0.2) -- (0.24,0.1) -- (0.54,0.1) -- (0.54,0.2) -- cycle;
\fill[line width=0.4pt,color=wwwwww,fill=wwwwww,fill opacity=0.1] (0.64,0.2) -- (0.64,0.1) -- (0.92,0.1) -- (0.92,0.2) -- cycle;
\fill[line width=0.4pt,dash pattern=on 1pt off 1pt,color=cxvqqq,fill=cxvqqq,fill opacity=0.15] (1,0.2) -- (1,0.1) -- (1.28,0.1) -- (1.28,0.2) -- cycle;
\fill[line width=0.4pt,color=wwwwww,fill=wwwwww,fill opacity=0.1] (1.4,0.2) -- (1.4,0.1) -- (1.68,0.1) -- (1.68,0.2) -- cycle;
\fill[line width=0.4pt,dash pattern=on 1pt off 1pt,color=cxvqqq,fill=cxvqqq,fill opacity=0.15] (1.76,0.2) -- (1.76,0.1) -- (2.04,0.1) -- (2.04,0.2) -- cycle;
\fill[line width=0.4pt,color=wwwwww,fill=wwwwww,fill opacity=0.1] (0.64,-0.2) -- (0.64,-0.3) -- (0.92,-0.3) -- (0.92,-0.2) -- cycle;
\fill[line width=0pt,color=cxvqqq,fill=cxvqqq,fill opacity=0.1] (0.97,-0.17) -- (0.97,-0.33) -- (2.03,-0.33) -- (2.03,-0.17) -- cycle;
\fill[line width=0.4pt,dash pattern=on 1pt off 1pt,color=cxvqqq,fill=cxvqqq,fill opacity=0.1] (1,-0.2) -- (1,-0.3) -- (1.28,-0.3) -- (1.28,-0.2) -- cycle;
\fill[line width=0.4pt,dash pattern=on 1pt off 1pt,color=cxvqqq,fill=cxvqqq,fill opacity=0.1] (1.36,-0.2) -- (1.36,-0.3) -- (1.64,-0.3) -- (1.64,-0.2) -- cycle;
\fill[line width=0.4pt,dash pattern=on 1pt off 1pt,color=cxvqqq,fill=cxvqqq,fill opacity=0.1] (1.72,-0.2) -- (1.72,-0.3) -- (2,-0.3) -- (2,-0.2) -- cycle;
\fill[line width=0.4pt,color=qqqqff,fill=qqqqff,fill opacity=0.1] (-0.14,-0.2) -- (-0.14,-0.3) -- (0.16,-0.3) -- (0.16,-0.2) -- cycle;
\fill[line width=0.4pt,color=qqqqff,fill=qqqqff,fill opacity=0.1] (0.24,-0.2) -- (0.24,-0.3) -- (0.54,-0.3) -- (0.54,-0.2) -- cycle;
\fill[line width=0pt,color=qqqqff,fill=qqqqff,fill opacity=0.1] (-0.17,-0.17) -- (-0.17,-0.33) -- (0.57,-0.33) -- (0.57,-0.17) -- cycle;
\draw [line width=0.4pt,color=qqwuqq] (0.24,0.2)-- (0.24,0.1);
\draw [line width=0.4pt,color=qqwuqq] (0.24,0.1)-- (0.54,0.1);
\draw [line width=0.4pt,color=qqwuqq] (0.54,0.1)-- (0.54,0.2);
\draw [line width=0.4pt,color=qqwuqq] (0.54,0.2)-- (0.24,0.2);
\draw [line width=0.4pt,color=wwwwww] (0.64,0.2)-- (0.64,0.1);
\draw [line width=0.4pt,color=wwwwww] (0.64,0.1)-- (0.92,0.1);
\draw [line width=0.4pt,color=wwwwww] (0.92,0.1)-- (0.92,0.2);
\draw [line width=0.4pt,color=wwwwww] (0.92,0.2)-- (0.64,0.2);
\draw [line width=0.4pt,dash pattern=on 1pt off 1pt,color=cxvqqq] (1,0.2)-- (1,0.1);
\draw [line width=0.4pt,dash pattern=on 1pt off 1pt,color=cxvqqq] (1,0.1)-- (1.28,0.1);
\draw [line width=0.4pt,dash pattern=on 1pt off 1pt,color=cxvqqq] (1.28,0.1)-- (1.28,0.2);
\draw [line width=0.4pt,dash pattern=on 1pt off 1pt,color=cxvqqq] (1.28,0.2)-- (1,0.2);
\draw [line width=0.4pt,color=wwwwww] (1.4,0.2)-- (1.4,0.1);
\draw [line width=0.4pt,color=wwwwww] (1.4,0.1)-- (1.68,0.1);
\draw [line width=0.4pt,color=wwwwww] (1.68,0.1)-- (1.68,0.2);
\draw [line width=0.4pt,color=wwwwww] (1.68,0.2)-- (1.4,0.2);
\draw [line width=0.4pt,dash pattern=on 1pt off 1pt,color=cxvqqq] (1.76,0.2)-- (1.76,0.1);
\draw [line width=0.4pt,dash pattern=on 1pt off 1pt,color=cxvqqq] (1.76,0.1)-- (2.04,0.1);
\draw [line width=0.4pt,dash pattern=on 1pt off 1pt,color=cxvqqq] (2.04,0.1)-- (2.04,0.2);
\draw [line width=0.4pt,dash pattern=on 1pt off 1pt,color=cxvqqq] (2.04,0.2)-- (1.76,0.2);
\draw [line width=0.4pt] (0.54,0.15)-- (0.64,0.15);
\draw [line width=0.4pt] (0.6024026386807567,0.15) -- (0.59,0.13346314842555734);
\draw [line width=0.4pt] (0.6024026386807567,0.15) -- (0.59,0.1665368515744427);
\draw [line width=0.4pt] (0.92,0.15)-- (1,0.15);
\draw [line width=0.4pt] (0.9724026386807567,0.15) -- (0.96,0.13346314842555734);
\draw [line width=0.4pt] (0.9724026386807567,0.15) -- (0.96,0.1665368515744427);
\draw [line width=0.4pt] (1.28,0.15)-- (1.4,0.15);
\draw [line width=0.4pt] (1.3524026386807568,0.15) -- (1.34,0.13346314842555734);
\draw [line width=0.4pt] (1.3524026386807568,0.15) -- (1.34,0.1665368515744427);
\draw [line width=0.8pt] (-0.14,-0.37)-- (2,-0.37);
\draw [line width=0.4pt,color=wwwwww] (0.64,-0.2)-- (0.64,-0.3);
\draw [line width=0.4pt,color=wwwwww] (0.64,-0.3)-- (0.92,-0.3);
\draw [line width=0.4pt,color=wwwwww] (0.92,-0.3)-- (0.92,-0.2);
\draw [line width=0.4pt,color=wwwwww] (0.92,-0.2)-- (0.64,-0.2);
\draw [line width=0.4pt] (0.92,-0.25)-- (0.97,-0.25);
\draw [line width=0.4pt] (0.9574026386807567,-0.25) -- (0.945,-0.2665368515744427);
\draw [line width=0.4pt] (0.9574026386807567,-0.25) -- (0.945,-0.2334631484255573);
\draw [line width=0.4pt,dash pattern=on 1pt off 1pt,color=cxvqqq] (1,-0.2)-- (1,-0.3);
\draw [line width=0.4pt,dash pattern=on 1pt off 1pt,color=cxvqqq] (1,-0.3)-- (1.28,-0.3);
\draw [line width=0.4pt,dash pattern=on 1pt off 1pt,color=cxvqqq] (1.28,-0.3)-- (1.28,-0.2);
\draw [line width=0.4pt,dash pattern=on 1pt off 1pt,color=cxvqqq] (1.28,-0.2)-- (1,-0.2);
\draw [line width=0.4pt,dash pattern=on 1pt off 1pt,color=cxvqqq] (1.36,-0.2)-- (1.36,-0.3);
\draw [line width=0.4pt,dash pattern=on 1pt off 1pt,color=cxvqqq] (1.36,-0.3)-- (1.64,-0.3);
\draw [line width=0.4pt,dash pattern=on 1pt off 1pt,color=cxvqqq] (1.64,-0.3)-- (1.64,-0.2);
\draw [line width=0.4pt,dash pattern=on 1pt off 1pt,color=cxvqqq] (1.64,-0.2)-- (1.36,-0.2);
\draw [line width=0.4pt,dash pattern=on 1pt off 1pt,color=cxvqqq] (1.72,-0.2)-- (1.72,-0.3);
\draw [line width=0.4pt,dash pattern=on 1pt off 1pt,color=cxvqqq] (1.72,-0.3)-- (2,-0.3);
\draw [line width=0.4pt,dash pattern=on 1pt off 1pt,color=cxvqqq] (2,-0.3)-- (2,-0.2);
\draw [line width=0.4pt,dash pattern=on 1pt off 1pt,color=cxvqqq] (2,-0.2)-- (1.72,-0.2);
\draw [line width=0.4pt,color=qqqqff] (-0.14,-0.2)-- (-0.14,-0.3);
\draw [line width=0.4pt,color=qqqqff] (-0.14,-0.3)-- (0.16,-0.3);
\draw [line width=0.4pt,color=qqqqff] (0.16,-0.3)-- (0.16,-0.2);
\draw [line width=0.4pt,color=qqqqff] (0.16,-0.2)-- (-0.14,-0.2);
\draw [line width=0.4pt,color=qqqqff] (0.24,-0.2)-- (0.24,-0.3);
\draw [line width=0.4pt,color=qqqqff] (0.24,-0.3)-- (0.54,-0.3);
\draw [line width=0.4pt,color=qqqqff] (0.54,-0.3)-- (0.54,-0.2);
\draw [line width=0.4pt,color=qqqqff] (0.54,-0.2)-- (0.24,-0.2);
\draw [line width=0.4pt] (0.54,-0.25)-- (0.64,-0.25);
\draw [line width=0.4pt] (0.6024026386807567,-0.25) -- (0.59,-0.2665368515744427);
\draw [line width=0.4pt] (0.6024026386807567,-0.25) -- (0.59,-0.2334631484255573);
\draw [line width=0.4pt] (0.39,-0.35)-- (0.39,-0.39);
\draw [line width=0.4pt] (1.14,-0.35)-- (1.14,-0.39);
\draw [line width=0.4pt] (1.9,0.2)-- (1.9,0.24);
\draw [line width=0.4pt] (1.9,0.24)-- (0.78,0.24);
\draw [line width=0.4pt] (0.78,0.24)-- (0.78,0.2);
\draw [line width=0.4pt] (0.78,0.207597361319168) -- (0.7634631484256577,0.22);
\draw [line width=0.4pt] (0.78,0.207597361319168) -- (0.7965368515743424,0.22);
\draw [line width=0.8pt] (0.24,0.06)-- (2,0.06);
\draw [line width=0.4pt] (0.39,0.08)-- (0.39,0.04);
\draw [line width=0.4pt] (1.14,0.08)-- (1.14,0.04);
\draw [line width=0.4pt] (1.9,0.08)-- (1.9,0.04);
\draw [line width=0.4pt] (1.86,-0.35)-- (1.86,-0.39);
\draw [line width=0.4pt] (0.01,-0.35)-- (0.01,-0.39);
\begin{scriptsize}
\draw [fill=black,shift={(2,-0.37)},rotate=270] (0,0) ++(0 pt,0.75pt) -- ++(0.649519052838329pt,-1.125pt)--++(-1.299038105676658pt,0 pt) -- ++(0.649519052838329pt,1.125pt);
\draw [fill=black,shift={(2,0.06)},rotate=270] (0,0) ++(0 pt,0.75pt) -- ++(0.649519052838329pt,-1.125pt)--++(-1.299038105676658pt,0 pt) -- ++(0.649519052838329pt,1.125pt);
\end{scriptsize}
\draw (0.63,0.205) node[anchor=north west] {\scriptsize{Surrogate}};
\draw (1.395,0.205) node[anchor=north west] {\scriptsize{Surrogate}};
\draw (0.63,-0.195) node[anchor=north west] {\scriptsize{Surrogate}};
\draw (0.99,0.34) node[anchor=north west] {\scriptsize{predicted output reused}};
\draw (-0.20,0.31) node[anchor=north west] {\color{qqwuqq}\scriptsize{rollout input: $x_{t}, p_{m}, p_{\mathrm{ref}}(t)$}};
\draw (0.99,-0.08) node[anchor=north west] {\color{cxvqqq}\scriptsize{forecast horizon ($H$ steps)}};
\draw (-0.2,-0.08) node[anchor=north west] {\color{qqqqff}\scriptsize{measurement window ($W$ steps)}};
\draw (1.04,0.21) node[anchor=north west] {\footnotesize{$\hat{x}_{t+1}$}};
\draw (1.04,-0.19) node[anchor=north west] {\footnotesize{$\hat{x}_{t+1}$}};
\draw (1.39,-0.19) node[anchor=north west] {\footnotesize{$\hat{x}_{t+2}$}};
\draw (1.74,-0.19) node[anchor=north west] {\footnotesize{$\hat{x}_{t+H}$}};
\draw (1.74,0.213) node[anchor=north west] {\footnotesize{$\hat{x}_{t+H_{\mathrm{roll}}}$}};
\draw (1.66,0.19) node[anchor=north west] {\footnotesize{\textbf{...}}};
\draw (1.62,-0.210) node[anchor=north west] {\footnotesize{\textbf{...}}};
\draw (0.14,-0.210) node[anchor=north west] {\footnotesize{\textbf{...}}};
\draw (-0.2,0.38) node[anchor=north west] {\footnotesize{\textbf{(i) Closed-loop simulation}}};
\draw (-0.2,-0.0) node[anchor=north west] {\footnotesize{\textbf{(ii) Measurement-window forecasting}}};
\draw (0.325,0.20) node[anchor=north west] {\footnotesize{$x_{t}$}};
\draw (0.325,-0.20) node[anchor=north west] {\footnotesize{$x_{t}$}};
\draw (-0.18,-0.37) node[anchor=north west] {\footnotesize{$x_{t-W+1}$}};
\draw (0.18,-0.39) node[anchor=north west] {\footnotesize{\textbf{...}}};
\draw (0.35,-0.37) node[anchor=north west] {\scriptsize{$t$}};
\draw (1.04,-0.37) node[anchor=north west] {\scriptsize{$t+1$}};
\draw (1.24,-0.38) node[anchor=north west] {\footnotesize{\textbf{...}}};
\draw (1.755,-0.37) node[anchor=north west] {\scriptsize{$t+H$}};
\draw (0.35,0.06) node[anchor=north west] {\scriptsize{$t$}};
\draw (1.04,0.06) node[anchor=north west] {\scriptsize{$t+1$}};
\draw (1.24,0.05) node[anchor=north west] {\footnotesize{\textbf{...}}};
\draw (1.74,0.06) node[anchor=north west] {\scriptsize{$t+H_{\mathrm{roll}}$}};
\draw (-0.15,-0.20) node[anchor=north west] {\footnotesize{$x_{t-W+1}$}};
\end{tikzpicture}
\vspace{-.6cm}
\caption{Task (i)~closed-loop simulation; (ii)~measurement-window forecasting, with $H$ outputs per forward pass and reported forecasting metrics based on $h{=}1$.}
\label{f2}
\end{figure}
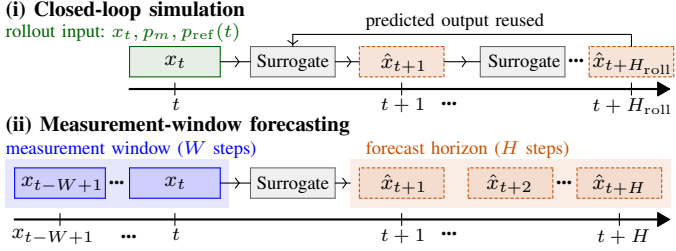

\subsection{Mamba Backbone and Task Conditioning}

The Mamba backbone models trajectory sequences with an input-modulated (selective) state-space recurrence. Let $h_t$ be the hidden representation at step $t$ and $s_t$ be the SSM internal state. The input-dependent recurrence is written as
\begin{equation}
\begin{aligned}
\Delta_t &= \mathrm{softplus}(h_tW_\Delta), \qquad
\bar{A}_t = \exp(\Delta_t\odot A),\\
s_t &= \bar{A}_t\odot s_{t-1}+\Delta_t\odot(h_tB),\\
z_t &= \sigma(h_tW_z)\odot\left(s_tC+D(h_t)\right).
\end{aligned}
\label{eq:mamba_update}
\end{equation}
Here, $W_\Delta$ and $W_z$ are learnable projection matrices, $\odot$ denotes element-wise multiplication, $\mathrm{softplus}(\cdot)$ and $\sigma(\cdot)$ denote the softplus and sigmoid activation functions, respectively, and $D(\cdot)$ is a learnable input projection. Let $A_{\log}$ be a learnable parameter and set $A=-\exp(A_{\log})$; therefore, $A<0$ and each component of $\bar{A}_t$ lies in $(0,1)$. In the present implementation, the input dependence of the SSM scan is introduced through $\Delta_t$, while $B$ and $C$ are static learned projections shared across the sequence. Here $h_t, z_t\in\mathbb{R}^{d_{\mathrm{model}}}$ and the SSM state $s_t\in\mathbb{R}^{d_s}$ (state dimension $d_s$); the recurrence is diagonal, so $A, \Delta_t, \bar{A}_t\in\mathbb{R}^{d_s}$ enter the state update element-wise, while $B\in\mathbb{R}^{d_{\mathrm{model}}\times d_s}$, $C\in\mathbb{R}^{d_s\times d_{\mathrm{model}}}$, $W_\Delta\in\mathbb{R}^{d_{\mathrm{model}}\times d_s}$, and $W_z\in\mathbb{R}^{d_{\mathrm{model}}\times d_{\mathrm{model}}}$ are the projections. Eq.~\eqref{eq:mamba_update} keeps the exact transition $\bar{A}_t=\exp(\Delta_t\!\odot\!A)$ and uses the first-order (Euler-type) input discretization $\Delta_t\!\odot\!(h_tB)$.

Task conditioning is implemented using FiLM~\cite{perez2018film}. The task embedding $c_m$ is passed through a multilayer perceptron to generate a scale term $g_{\mathrm{FiLM}}$ and a shift term $\beta$, which modulate the layer normalization in each Mamba block:
\begin{equation}
\gamma=\mathbf{1}+g_{\mathrm{FiLM}},\qquad
\mathrm{LN}_{\mathrm{FiLM}}(u)=\gamma\odot \mathrm{LN}(u)+\beta.
\label{eq:film}
\end{equation}
$\mathrm{LN}(\cdot)$ denotes standard layer normalization, $u$ the normalized input, and $\mathbf{1}$ is an all-ones vector. At initialization, $g_{\mathrm{FiLM}}=0$ and $\beta=0$, so FiLM reduces to standard layer normalization.

\subsection{MoE Routing and Expert Mixing}

The MoE routing layer introduces task- and operating-condition-dependent adaptation after the shared Mamba representation. Let $h_r$ denote the hidden representation used for routing. In the simulation task, $h_r$ corresponds to the hidden representation at the current rollout step; in the forecasting task, $h_r$ corresponds to the final hidden representation of the measurement window. The router maps the concatenation of $h_r$ and the task embedding $c_m$ to expert logits:
\begin{equation}
\ell=r([h_r;c_m])\in\mathbb{R}^{K},
\label{eq:router_logits}
\end{equation}
where $r(\cdot)$ is the routing network, $K$ is the number of experts, and $[\,;\,]$ denotes vector concatenation. The expert weights are obtained by softmax normalization:
\begin{equation}
g_k=\frac{\exp(\ell_k)}{\sum_{q=1}^{K}\exp(\ell_q)},\qquad
k=1,\ldots,K.
\label{eq:router}
\end{equation}

Let $E_k(\cdot)$ be the $k$th expert subnetwork. The expert-mixed output
\begin{equation}
e=\sum_{k=1}^{K}g_kE_k(h_r),\qquad e\in\mathbb{R}^{d_{\mathrm{model}}}.
\label{eq:moe_mix}
\end{equation}
This paper uses dense soft routing, in which all experts participate in each forward pass with soft weights. For the model scale considered here, this design has a small computational overhead and avoids the load-imbalance issue that may arise in sparse top-$k$ routing. The expert combination is thus jointly determined by the current dynamic features and the simulation or forecasting task identity.

\subsection{Unified Output Head and Task-Specific Readout}

Given the expert-mixed output $e$ (the routed representation at the current step), a shared output Mamba block $G_{\mathrm{out}}(\cdot)$ produces the output representation $h_o\in\mathbb{R}^{d_{\mathrm{model}}}$, and a unified linear output head generates an $H$-output residual matrix:
\begin{equation}
h_o=G_{\mathrm{out}}(e),\quad
o=\mathrm{Head}(h_o)=[r_1,\ldots,r_H],
\label{eq:head}
\end{equation}
where $r_h\in\mathbb{R}^{n_x}$ is the $h$-th residual output (row of $o$). $G_{\mathrm{out}}$ is applied one step at a time: in simulation it carries its SSM state across the rollout, whereas for forecasting it is applied once to the final-window representation, and the $H$ residuals are produced by the linear head rather than by $G_{\mathrm{out}}$. The output space has the same dimension as the current-output vector $x_t$, so the same output head can generate output residuals for both tasks. The two tasks share the same residual matrix but read it differently. The simulation task uses only the first-step residual $r_{1,t}$ at each autoregressive step:
\begin{equation}
\hat{x}^{\mathrm{sim}}_{t+1}=\hat{x}^{\mathrm{sim}}_{t}+r_{1,t}.
\label{eq:sim_head}
\end{equation}
The forecasting task uses all $H$ residuals and adds them to the last observed output $x_t$ of the measurement window:
\begin{equation}
\hat{x}^{\mathrm{fcast}}_{t+h}=x_t+r_h,\qquad h=1,\ldots,H.
\label{eq:fcast_head}
\end{equation}
Therefore, the unified output head defines a common output-residual space, while task differences are handled by the input structure, FiLM conditioning, MoE routing, and task-specific training objectives.

\section{Task-Matched Training Objectives}
\label{sec:training}

This section defines the training objectives for the unified Mamba--MoE surrogate. Training has two stages. The Mamba backbone is first pretrained with one-step output residual prediction to learn a shared current-response representation. The complete model is then jointly trained so that simulation and forecasting follow their respective prediction forms.

\subsection{Backbone Pretraining}

The pretraining stage forms a one-step residual target from output trajectories. Let $h_W$ denote the final hidden representation produced by the Mamba backbone from an output window of length $W$. A temporary residual head $H_{\mathrm{PT}}(\cdot)$ predicts $\delta x_t=x_{t+1}-x_t$. The pretraining loss is
\begin{equation}
\mathcal{L}_{\mathrm{PT}}
=
\frac{1}{N_B n_x}
\sum_{i=1}^{N_B}
\sum_{j=1}^{n_x}
\left(
H_{\mathrm{PT}}(h_W^{(i)})_j
-
\delta x^{(i)}_j
\right)^2 ,
\label{eq:lpt}
\end{equation}
where $N_B$ is the batch size. The task-conditioning vector is set to zero during pretraining, so the backbone first learns a task-agnostic transition representation. After pretraining, $H_{\mathrm{PT}}$ is discarded.

\subsection{Task-Matched Losses}

Both tasks predict the same current outputs, but their losses match different prediction forms. The simulation task uses autoregressive rollout. Since each predicted output is fed back as the next input, its loss is computed over $H_{\mathrm{roll}}$ rollout steps:
\begin{equation}
\mathcal{L}_{\mathrm{sim}}
=
\frac{1}{H_{\mathrm{roll}}N_B n_x}
\sum_{h=1}^{H_{\mathrm{roll}}}
\sum_{i=1}^{N_B}
\sum_{j=1}^{n_x}
\left(
\hat{x}^{(i),\mathrm{sim}}_{t+h,j}
-
x^{(i)}_{t+h,j}
\right)^2 ,
\label{eq:lsim}
\end{equation}
where $H_{\mathrm{roll}}$ is the simulation rollout horizon used in training. The forecasting task directly maps a recent measurement window to $H$ future output residuals. Its loss supervises the full forecasting horizon:
\begin{equation}
\mathcal{L}_{\mathrm{fcast}}
=
\frac{1}{N_B H n_x}
\sum_{i=1}^{N_B}
\sum_{h=1}^{H}
\sum_{j=1}^{n_x}
\left(
\hat{x}^{(i),\mathrm{fcast}}_{t+h,j}
-
x^{(i)}_{t+h,j}
\right)^2 .
\label{eq:lforec}
\end{equation}

\subsection{Joint Objective and Algorithm}

Joint training optimizes both task losses and two lightweight MoE regularizers. Let $\bar{g}_k$ be the average soft routing weight of expert $k$ in a batch, and let $\bar{a}_k$ be its top-1 assignment frequency; both statistics are pooled over the simulation and forecasting routing decisions in the joint batch, since the router is shared by both tasks. The load-balancing term is
\begin{equation}
\mathcal{L}_{\mathrm{LB}}
=
K\sum_{k=1}^{K}
\bar{g}_k\,\mathrm{sg}(\bar{a}_k),
\label{eq:llb}
\end{equation}
where $K$ is the number of experts and $\mathrm{sg}(\cdot)$ denotes stop-gradient. The expert-diversity term penalizes positive cosine similarity between different expert outputs:
\begin{equation}
\mathcal{L}_{\mathrm{div}}
=
\frac{2}{K(K-1)}
\sum_{k<l}
\left[
\cos(E_k(h_r),E_l(h_r))
\right]_+
\label{eq:ldiv}
\end{equation}
where $[a]_+=\max\{a,0\}$. For $K<2$ the diversity term $\mathcal{L}_{\mathrm{div}}$ is set to zero and the load-balancing term $\mathcal{L}_{\mathrm{LB}}$ is omitted, so the $K=1$ ablation is trained without MoE regularization. The total objective is
\begin{equation}
\mathcal{L}_{\mathrm{total}}
=
\lambda_{\mathrm{sim}}\mathcal{L}_{\mathrm{sim}}
+
\lambda_{\mathrm{fcast}}\mathcal{L}_{\mathrm{fcast}}
+
\lambda_{\mathrm{LB}}\mathcal{L}_{\mathrm{LB}}
+
\lambda_{\mathrm{div}}\mathcal{L}_{\mathrm{div}},
\label{eq:ltotal}
\end{equation}
where all $\lambda$ values are non-negative weights specified in the experimental setup. Algorithm~\ref{alg:train} summarizes the two-stage training procedure.

\begin{algorithm}[!t]
\small
\caption{Unified Mamba--MoE Training}
\label{alg:train}
\begin{algorithmic}[1]
\Require $\mathcal{D}_{\mathrm{PT}}$, $\mathcal{D}_{\mathrm{sim}}$,
$\mathcal{D}_{\mathrm{fcast}}$, $W$, $H$, $H_{\mathrm{roll}}$,
$N_{\mathrm{PT}}$, $N_{\mathrm{joint}}$, $\{\lambda_{\cdot}\}$
\Ensure Trained parameters $\theta$
\State Initialize $\theta$ and set the task-conditioning vector to $\mathbf{0}$.
\For{$e=1,\ldots,N_{\mathrm{PT}}$}
\State Sample $(x_{t-W+1:t},\delta x_t)\sim\mathcal{D}_{\mathrm{PT}}$.
\State Update input projection, backbone, and $H_{\mathrm{PT}}$ using $\mathcal{L}_{\mathrm{PT}}$.
\EndFor
\State Enable $c_m$, FiLM, MoE routing, and the shared output head.
\For{$e=1,\ldots,N_{\mathrm{joint}}$}
\State Sample $B_{\mathrm{sim}}\sim\mathcal{D}_{\mathrm{sim}}$, compute
$\mathcal{L}_{\mathrm{sim}}$, and record its routing weights.
\State Sample $B_{\mathrm{fcast}}\sim\mathcal{D}_{\mathrm{fcast}}$, compute
$\mathcal{L}_{\mathrm{fcast}}$, and record its routing weights.
\State Form $\mathcal{L}_{\mathrm{LB}}$ from the pooled routing weights
and $\mathcal{L}_{\mathrm{div}}$ from the corresponding expert outputs.
\State Update $\theta$ with AdamW to minimize
$\mathcal{L}_{\mathrm{total}}$.
\EndFor
\end{algorithmic}
\end{algorithm}

\begin{figure}[!t]
\centering
\includegraphics[trim={0 1.10cm 0 1.14cm}, clip, width=\columnwidth]{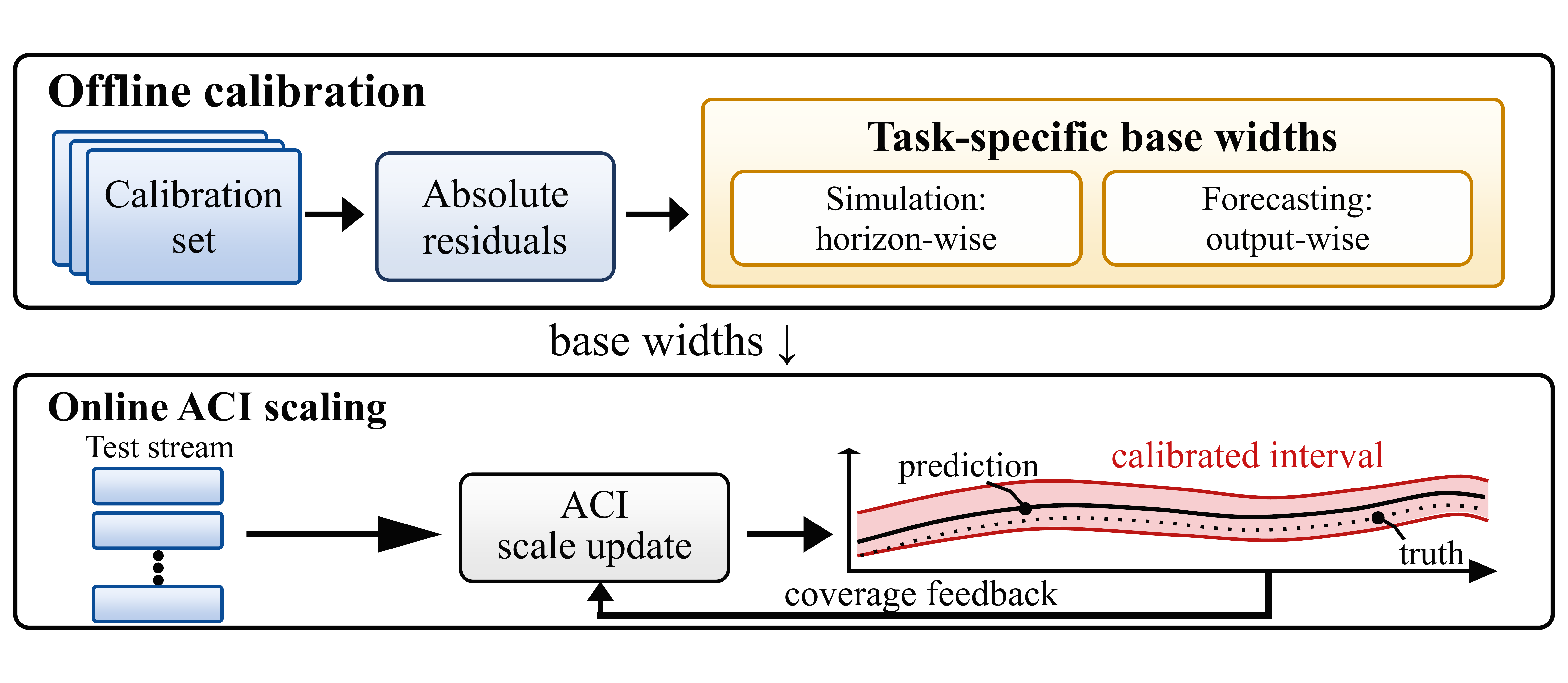}
\vspace{-.7cm}
\caption{Adaptive conformal calibration workflow.}
\label{fig:aci}
\end{figure}

\section{Adaptive Prediction-Interval Calibration}
\label{sec:aci}

Fig.~\ref{fig:aci} summarizes the adaptive conformal inference (ACI) workflow: task-specific conformal widths are computed on the calibration split, and a per-output scale factor is adapted online to track the target coverage level.

\subsection{Calibration Challenges for Simulation and Forecasting}

Point predictions alone do not indicate whether a transient estimate is reliable. Standard conformal prediction (CP)~\cite{angelopoulos2021gentle} provides distribution-free coverage under exchangeable calibration and test residuals, but the two prediction modes considered here have different residual structures. Simulation residuals are horizon-dependent because rollout reuses predictions, whereas forecasting residuals are more homogeneous after closest-window merging; residual levels may still shift when the test operating point differs from the calibration set. Because the residuals come from overlapping temporal windows, the coverage reported in \S\ref{sec:results} is empirical rather than a finite-sample distribution-free guarantee.

\subsection{Task-Specific Conformal Base Widths}

Simulation uses horizon-wise conformal widths because rollout uncertainty grows with horizon. For target miscoverage $\alpha$, horizon $h$, and output dimension $j$, let
\[
\epsilon^{\mathrm{sim}}_{(1),h,j}
\le \cdots \le
\epsilon^{\mathrm{sim}}_{(n_{h,j}),h,j}
\]
denote the sorted absolute calibration residuals, where $n_{h,j}$ is the corresponding calibration count. The split-conformal rank
\begin{equation}
\rho^{\mathrm{sim}}_{h,j}
=
\min\left\{
\left\lceil (n_{h,j}+1)(1-\alpha)\right\rceil,\,
n_{h,j}
\right\},
\label{eq:sim_rank}
\end{equation}
and the simulation base width is
\begin{equation}
q^{\mathrm{sim}}_{h,j}
=
\epsilon^{\mathrm{sim}}_{(\rho^{\mathrm{sim}}_{h,j}),h,j}.
\label{eq:qbase_sim}
\end{equation}

Forecasting is calibrated after closest-window merging. For target time $t'$, the eligible forecast origins are
\begin{equation}
\mathcal{A}_i(t')
=
\{t\in\mathcal{W}_i:\ 1\le t'-t\le H\},
\label{eq:merge_window_set}
\end{equation}
where $\mathcal{W}_i$ contains the last observation time of each input window in trajectory $i$. We select the most recent eligible origin and its associated horizon:
\begin{equation}
t_i^\star(t')=\max \mathcal{A}_i(t'),
\qquad
h_i^\star(t')=t'-t_i^\star(t').
\label{eq:closest_window}
\end{equation}
The merged prediction is
\begin{equation}
\hat{x}^{\mathrm{merge}}_{i,t',j}
=
\hat{x}^{\mathrm{fcast}}_{i,t_i^\star(t'),h_i^\star(t'),j}.
\label{eq:closest_merge}
\end{equation}
Forecasting residuals are then computed as
\begin{equation}
\epsilon^{\mathrm{fcast}}_{i,t',j}
=
\left|x_{i,t',j}-\hat{x}^{\mathrm{merge}}_{i,t',j}\right|.
\label{eq:fcast_residual}
\end{equation}
By sorting the $n_j$ merged residuals for output dimension $j$ as
\[
\epsilon^{\mathrm{fcast}}_{(1),j}
\le \cdots \le
\epsilon^{\mathrm{fcast}}_{(n_j),j},
\]
the forecasting base width is
\begin{equation}
\begin{aligned}
\rho^{\mathrm{fcast}}_j
&=
\min\left\{
\left\lceil (n_j+1)(1-\alpha)\right\rceil,\,
n_j
\right\},\\
q^{\mathrm{fcast}}_j
&=
\epsilon^{\mathrm{fcast}}_{(\rho^{\mathrm{fcast}}_j),j}.
\end{aligned}
\label{eq:qbase_fcast}
\end{equation}
All base widths use the calibration split only.

\subsection{Online ACI Scaling}

Operating-point mismatch between calibration and test trajectories can shift the residual level. ACI therefore scales the conformal base widths using sequential coverage feedback~\cite{gibbs2021adaptive}. Let $u$ index the evaluated intervals and let $q_{u,j}$ denote the corresponding base width: $q^{\mathrm{sim}}_{h,j}$ in simulation and $q^{\mathrm{fcast}}_j$ in forecasting. For each output dimension $j$, the interval
\begin{equation}
\hat{C}_{u,j}
=
\left[
\hat{x}_{u,j}
\pm
\kappa_u(j)q_{u,j}
\right],
\qquad
\kappa_u(j)=\exp(z_u(j)).
\label{eq:aci_interval}
\end{equation}

The log-scale update is
\begin{equation}
z_{u+1}(j)
=
\Pi_{[-c,c]}
\left(
z_u(j)
+
\eta_{\mathrm{ACI}}
\left(
\mathbf{1}\!\left[x_{u,j}\notin \hat{C}_{u,j}\right]
-
\alpha
\right)
\right),
\label{eq:aci_update}
\end{equation}
with
\begin{equation}
\Pi_{[-c,c]}(z)
=
\min\{\max\{z,-c\},c\}.
\label{eq:clip_operator}
\end{equation}
We use $c=5$. The validation stream is processed sequentially from $z_1(j)=0$; each test trajectory is reset to the same final state obtained from this stream. The update is strictly prequential: $\hat{C}_{u,j}$ is formed before $x_{u,j}$ is revealed, and the realized label is used only to score coverage and update $z_{u+1}(j)$; thus no future test label enters interval construction. Each merged forecast target is processed when its label becomes available.

\section{Case Studies and Results}
\label{sec:results}

We evaluate the unified model from six perspectives: clean-condition accuracy and model size, ablation behavior, robustness under operating-point shift and measurement corruption, cross-topology forecasting and adaptation on a 9-bus system, cross-system CHIL adaptation, and interval reliability after ACI calibration.\footnote{\href{https://github.com/hgwang135/Mamba-MOE}{Data and code} will be available upon publication.}

\subsection{Experimental Setup and Baselines}

The baseline dataset comprises 150 offline transient trajectories generated by integrating the inverter--infinite-bus model in Section~III-A using MATLAB's \texttt{ode15s}. The solver uses adaptive internal steps, and outputs are recorded every 5\,ms. We split the trajectories into 80 training, 20 validation, and 50 test trajectories, and apply $z$-score normalization to all output channels before training. Each trajectory contains a step change in the active-power reference from $0.5$\,pu to $p_m$ at $t=3$\,s. The trajectory-level power command varies across cases as $p_m \sim \mathcal{U}(0.4, 0.8)$\,pu, while the reactive-power reference is held fixed at its system setpoint. The analysis window covers $t \in [2, 4]$\,s; at this recording interval, each trajectory contains $L = 401$ samples. For Tables~I--III, we report root-mean-square error (RMSE) and its global-range-normalized form (G-NRMSE):
\begin{equation}
\mathrm{G\text{-}NRMSE}_j(\%)
=
\frac{\mathrm{RMSE}_j}{\mathrm{Range}^{\mathrm{train}}_j}
\times 100 ,
\label{eq:gnrmse}
\end{equation}
where $j = 1,\ldots,n_x$ indexes the output dimension and $\mathrm{Range}^\mathrm{train}_j = \max(x^{\mathrm{train}}_j) - \min(x^{\mathrm{train}}_j)$ is the per-dimension dynamic range observed during training. The per-dimension values are $\mathrm{Range}^{\mathrm{train}}_{i^{\mathrm{cv}}_d}=0.783$\,pu, $\mathrm{Range}^{\mathrm{train}}_{i^{\mathrm{cv}}_q}=0.745$\,pu, $\mathrm{Range}^{\mathrm{train}}_{i_{r}^{\mathrm{filt}}}= 0.783$\,pu, and $\mathrm{Range}^{\mathrm{train}}_{i_{i}^{\mathrm{filt}}}= 0.745$\,pu. Because all four ranges are comparable, cross-channel G-NRMSE differences are not caused by unequal denominators. In absolute terms, a G-NRMSE near $0.1\%$ corresponds to a current RMSE on the order of $10^{-3}$\,pu, i.e., less than one percent of the per-unit current range used for normalization, so the reported errors are small at the system level.

For interval reliability, let $m\in\{\mathrm{sim},\mathrm{fcast}\}$ denote the task and let $\mathcal{I}_m$ be its evaluation set. For simulation, $\mathcal{I}_{\mathrm{sim}}$ contains rollout evaluation points. For forecasting, $\mathcal{I}_{\mathrm{fcast}}$ contains target-time predictions after closest-window merging. For $u\in\mathcal{I}_m$, let $\hat{C}_{u,j}$ denote the prediction interval for output $j$. We report marginal coverage (MCOV) and mean prediction interval width (MPIW) as
\begin{equation}
\mathrm{MCOV}^{(m)}
=
\frac{1}{|\mathcal{I}_m|n_x}
\sum_{u\in\mathcal{I}_m}
\sum_{j=1}^{n_x}
\mathbf{1}\!\left[x_{u,j}\in \hat{C}_{u,j}\right],
\label{eq:mcov_task}
\end{equation}
\begin{equation}
\mathrm{MPIW}^{(m)}
=
\frac{1}{|\mathcal{I}_m|n_x}
\sum_{u\in\mathcal{I}_m}
\sum_{j=1}^{n_x}
\mathrm{width}\!\left(\hat{C}_{u,j}\right).
\label{eq:mpiw_task}
\end{equation}

Static CP keeps the conformal scale fixed at $\kappa_u(j)=1$, whereas ACI updates the scale factors online. The target coverage is $1-\alpha=95\%$. Unless otherwise stated, the unified model uses $d_{\mathrm{model}}=128$, four Mamba blocks, $K=4$ experts, $d_c=32$, $d_s=16$, $W=10$, $H=24$, and $H_{\mathrm{roll}}=20$. Training supervises all $H$ forecast outputs, but unit-stride origins and closest-window merging select $h=1$, so the reported forecasting errors and intervals are 5-ms-ahead. We use $\lambda_{\mathrm{sim}}=\lambda_{\mathrm{fcast}}=1$ and $\lambda_{\mathrm{LB}}= \lambda_{\mathrm{div}}= 5\times10^{-4}$; AdamW starts at $10^{-3}$ for pretraining and $5\times10^{-4}$ for joint training, with weight decay $10^{-5}$; and ACI uses $\alpha=0.05$, $\eta_{\mathrm{ACI}}=0.02$, and $c=5$. DeepONet, Transformer, and Mamba are each trained as separate simulation and forecasting specialists with task-matched objectives; Table~\ref{tab:accuracy} reports the combined parameter count of each pair. \emph{DeepONet}~\cite{moya2023deeponet, sun2023deepgraphonet, lin2023learning, moya2023dae, moya2023bayesian} learns the operator mapping from input functions to output trajectories. \emph{Transformer}~\cite{vaswani2017attention} uses a standard encoder--decoder with positional encoding and causal masking for autoregressive rollout. The Mamba specialists use the same selective state-space block design as the Mamba--MoE backbone but exclude FiLM conditioning and MoE routing, making them the closest architectural references. Training schedules are architecture- and task-specific, while all methods use the same data split and evaluation protocol.

\subsection{Clean-Condition Comparison with Specialists}

Table~\ref{tab:accuracy} compares clean-condition G-NRMSE and RMSE. The zero-delta baseline sets the output increment to zero at every prediction step and tests whether a learned model improves on a persistence predictor. In simulation, Mamba--MoE reduces the $d/r$ output errors from about $10\%$ for zero-delta to at most $0.21\%$, and the $q/i$ output errors from $2.72\%$ to about $1.39\%$. In forecasting, Mamba--MoE reduces $i^{\mathrm{cv}}_d$ and $i_{r}^{\mathrm{filt}}$ from $0.81\%$ to $0.10\%$ and $0.12\%$, respectively. The forecasting errors for $i^{\mathrm{cv}}_q$ and $i_{i}^{\mathrm{filt}}$ remain close to zero-delta because these $q/i$ outputs remain near zero in steady state, so brief transient excursions define the range and leave limited G-NRMSE headroom. DeepONet and Transformer offer additional architectural references, but neither consistently outperforms Mamba--MoE across both tasks. The Mamba specialist pair provides the strongest clean-condition specialist reference, especially for forecasting $i^{\mathrm{cv}}_d$ and $i_{r}^{\mathrm{filt}}$. Mamba--MoE does not exceed every specialist metric under clean conditions; instead, it keeps both tasks in the same low-error regime while using a single unified model. At 800,260 parameters versus 921,316 for the Mamba specialist pair, the unified model uses about $13\%$ fewer parameters.

\begin{table*}[t]
\renewcommand{\arraystretch}{1.05}
\caption{Clean-condition accuracy and parameter counts; baseline
counts sum the parameters of the simulation and forecasting specialists.}
\label{tab:accuracy}
\centering
\scriptsize
\setlength{\tabcolsep}{4pt}
\begin{tabular}{l r l c c c c}
\toprule
\textbf{Model} & \textbf{Params} & \textbf{Output}
& \textbf{G-NRMSE Simulation (\%)} & \textbf{RMSE Simulation (pu)}
& \textbf{G-NRMSE Forecasting (\%)} & \textbf{RMSE Forecasting (pu)} \\
\midrule
\multirow{4}{*}{Zero-delta} &
\multirow{4}{*}{---} &
$i^{\mathrm{cv}}_{d}$ & 10.17 & $7.97\times10^{-2}$ & 0.81 & $6.38\times10^{-3}$ \\
& & $i^{\mathrm{cv}}_{q}$ & 2.72 & $2.02\times10^{-2}$ & 1.56 & $1.16\times10^{-2}$ \\
& & $i_{r}^{\mathrm{filt}}$ & 10.18 & $7.97\times10^{-2}$ & 0.81 & $6.38\times10^{-3}$ \\
& & $i_{i}^{\mathrm{filt}}$ & 2.72 & $2.03\times10^{-2}$ & 1.56 & $1.16\times10^{-2}$ \\
\midrule
\multirow{4}{*}{DeepONet} &
\multirow{4}{*}{704,648} &
$i^{\mathrm{cv}}_{d}$ & 0.19 & $1.53\times10^{-3}$ & 0.10 & $7.59\times10^{-4}$ \\
& & $i^{\mathrm{cv}}_{q}$ & 1.48 & $1.10\times10^{-2}$ & 1.42 & $1.06\times10^{-2}$ \\
& & $i_{r}^{\mathrm{filt}}$ & 0.19 & $1.49\times10^{-3}$ & 0.09 & $7.21\times10^{-4}$ \\
& & $i_{i}^{\mathrm{filt}}$ & 1.47 & $1.10\times10^{-2}$ & 1.42 & $1.05\times10^{-2}$ \\
\midrule
\multirow{4}{*}{Transformer} &
\multirow{4}{*}{1,107,300} &
$i^{\mathrm{cv}}_{d}$ & 0.42 & $3.29\times10^{-3}$ & 0.06 & $4.44\times10^{-4}$ \\
& & $i^{\mathrm{cv}}_{q}$ & 1.42 & $1.06\times10^{-2}$ & 1.41 & $1.05\times10^{-2}$ \\
& & $i_{r}^{\mathrm{filt}}$ & 0.43 & $3.34\times10^{-3}$ & 0.05 & $3.79\times10^{-4}$ \\
& & $i_{i}^{\mathrm{filt}}$ & 1.43 & $1.06\times10^{-2}$ & 1.41 & $1.05\times10^{-2}$ \\
\midrule
\multirow{4}{*}{\makecell[l]{Mamba\\(specialist pair)}} &
\multirow{4}{*}{921,316} &
$i^{\mathrm{cv}}_{d}$ & 0.12 & $9.48\times10^{-4}$ & 0.02 & $1.87\times10^{-4}$ \\
& & $i^{\mathrm{cv}}_{q}$ & 1.39 & $1.03\times10^{-2}$ & 1.41 & $1.05\times10^{-2}$ \\
& & $i_{r}^{\mathrm{filt}}$ & 0.11 & $8.78\times10^{-4}$ & 0.02 & $1.62\times10^{-4}$ \\
& & $i_{i}^{\mathrm{filt}}$ & 1.39 & $1.03\times10^{-2}$ & 1.41 & $1.05\times10^{-2}$ \\
\midrule
\multirow{4}{*}{\makecell[l]{Mamba--MoE\\(unified)}} &
\multirow{4}{*}{800,260} &
$i^{\mathrm{cv}}_{d}$ & 0.14 & $1.06\times10^{-3}$ & 0.10 & $7.93\times10^{-4}$ \\
& & $i^{\mathrm{cv}}_{q}$ & 1.39 & $1.04\times10^{-2}$ & 1.41 & $1.05\times10^{-2}$ \\
& & $i_{r}^{\mathrm{filt}}$ & 0.21 & $1.64\times10^{-3}$ & 0.12 & $9.70\times10^{-4}$ \\
& & $i_{i}^{\mathrm{filt}}$ & 1.39 & $1.04\times10^{-2}$ & 1.40 & $1.05\times10^{-2}$ \\
\bottomrule
\end{tabular}
\end{table*}

\subsection{Ablation Analysis}

Table~\ref{tab:ablation} examines four design choices. First, two task-specific heads reduce forecasting G-NRMSE for the $d/r$ outputs from $0.10\%/0.12\%$ to $0.03\%/0.03\%$, while changing simulation errors from $0.14\%/0.21\%$ to $0.15\%/0.16\%$. Neither head configuration dominates all tasks and outputs. We retain the shared head because both tasks predict residuals in the same normalized output space.

Second, compared with $K=4$, $K=1$ gives lower error only for simulation $i_r^{\mathrm{filt}}$; all remaining metrics are higher or essentially unchanged. Thus, $K=4$ gives lower errors for more outputs, but not for every output. Table~\ref{tab:gen} shows greater routing benefits as operating points move farther from the training range.

Third, removing FiLM lowers simulation G-NRMSE for the $d/r$ outputs from $0.14\%/0.21\%$ to $0.13\%/0.15\%$, but raises forecasting errors from $0.10\%/0.12\%$ to $0.13\%/0.17\%$. Its effect is task-dependent and mainly benefits forecasting $d/r$ outputs. Because the router still uses the task embedding $c_m$, this ablation removes FiLM but not task conditioning entirely.

Finally, removing either supervised loss increases error for that task. With $\lambda_{\mathrm{fcast}}=0$, forecasting G-NRMSE for the $d/r$ outputs rises to $18.70\%/22.74\%$; with $\lambda_{\mathrm{sim}}=0$, simulation errors rise to $65.65\%/121.70\%$. Both losses are required for the unified surrogate under the current training objective.

\begin{table*}[t]
\renewcommand{\arraystretch}{1.05}
\caption{Ablation study on design components.}
\label{tab:ablation}
\centering
\scriptsize
\setlength{\tabcolsep}{4pt}
\begin{tabular}{l c c c c c l c c c c}
\toprule
\textbf{Variant} & \textbf{Head} & $K$ & \textbf{FiLM}
& $\lambda_{\mathrm{sim}}$ & $\lambda_{\mathrm{fcast}}$ & \textbf{Output}
& \textbf{G-NRMSE Simulation} & \textbf{RMSE Simulation} & \textbf{G-NRMSE Forecasting} & \textbf{RMSE Forecasting} \\
& & & & & & & \textbf{(\%)} & \textbf{(pu)} & \textbf{(\%)} & \textbf{(pu)} \\
\midrule
\multirow{4}{*}{\makecell[l]{Mamba--MoE\\Baseline}} &
\multirow{4}{*}{1H} & \multirow{4}{*}{4} & \multirow{4}{*}{$\checkmark$} &
\multirow{4}{*}{1.0} & \multirow{4}{*}{1.0} &
$i^{\mathrm{cv}}_d$ & 0.14 & $1.06\times10^{-3}$ & 0.10 & $7.93\times10^{-4}$ \\
& & & & & & $i^{\mathrm{cv}}_q$ & 1.39 & $1.04\times10^{-2}$ & 1.41 & $1.05\times10^{-2}$ \\
& & & & & & $i_{r}^{\mathrm{filt}}$ & 0.21 & $1.64\times10^{-3}$ & 0.12 & $9.70\times10^{-4}$ \\
& & & & & & $i_{i}^{\mathrm{filt}}$ & 1.39 & $1.04\times10^{-2}$ & 1.40 & $1.05\times10^{-2}$ \\
\midrule
\multirow{4}{*}{\makecell[l]{(A) Two-head}} &
\multirow{4}{*}{2H} & \multirow{4}{*}{4} & \multirow{4}{*}{$\checkmark$} &
\multirow{4}{*}{1.0} & \multirow{4}{*}{1.0} &
$i^{\mathrm{cv}}_d$ & 0.15 & $1.15\times10^{-3}$ & 0.03 & $2.60\times10^{-4}$ \\
& & & & & & $i^{\mathrm{cv}}_q$ & 1.39 & $1.04\times10^{-2}$ & 1.41 & $1.05\times10^{-2}$ \\
& & & & & & $i_{r}^{\mathrm{filt}}$ & 0.16 & $1.23\times10^{-3}$ & 0.03 & $2.59\times10^{-4}$ \\
& & & & & & $i_{i}^{\mathrm{filt}}$ & 1.39 & $1.04\times10^{-2}$ & 1.41 & $1.05\times10^{-2}$ \\
\midrule
\multirow{4}{*}{\makecell[l]{(B) $K=1$\\(no routing)}} &
\multirow{4}{*}{1H} & \multirow{4}{*}{1} & \multirow{4}{*}{$\checkmark$} &
\multirow{4}{*}{1.0} & \multirow{4}{*}{1.0} &
$i^{\mathrm{cv}}_d$ & 0.15 & $1.20\times10^{-3}$ & 0.14 & $1.06\times10^{-3}$ \\
& & & & & & $i^{\mathrm{cv}}_q$ & 1.39 & $1.04\times10^{-2}$ & 1.41 & $1.05\times10^{-2}$ \\
& & & & & & $i_{r}^{\mathrm{filt}}$ & 0.14 & $1.10\times10^{-3}$ & 0.15 & $1.17\times10^{-3}$ \\
& & & & & & $i_{i}^{\mathrm{filt}}$ & 1.39 & $1.04\times10^{-2}$ & 1.41 & $1.05\times10^{-2}$ \\
\midrule
\multirow{4}{*}{\makecell[l]{(C) w/o FiLM}} &
\multirow{4}{*}{1H} & \multirow{4}{*}{4} & \multirow{4}{*}{$\times$} &
\multirow{4}{*}{1.0} & \multirow{4}{*}{1.0} &
$i^{\mathrm{cv}}_d$ & 0.13 & $1.04\times10^{-3}$ & 0.13 & $1.02\times10^{-3}$ \\
& & & & & & $i^{\mathrm{cv}}_q$ & 1.39 & $1.04\times10^{-2}$ & 1.41 & $1.05\times10^{-2}$ \\
& & & & & & $i_{r}^{\mathrm{filt}}$ & 0.15 & $1.18\times10^{-3}$ & 0.17 & $1.30\times10^{-3}$ \\
& & & & & & $i_{i}^{\mathrm{filt}}$ & 1.39 & $1.03\times10^{-2}$ & 1.41 & $1.05\times10^{-2}$ \\
\midrule
\multirow{4}{*}{\makecell[l]{(D1) $\lambda_{\mathrm{fcast}}{=}0$}} &
\multirow{4}{*}{1H} & \multirow{4}{*}{4} & \multirow{4}{*}{$\checkmark$} &
\multirow{4}{*}{1.0} & \multirow{4}{*}{0} &
$i^{\mathrm{cv}}_d$ & 0.10 & $8.10\times10^{-4}$ & 18.70 & $1.46\times10^{-1}$ \\
& & & & & & $i^{\mathrm{cv}}_q$ & 1.39 & $1.04\times10^{-2}$ & 14.94 & $1.11\times10^{-1}$ \\
& & & & & & $i_{r}^{\mathrm{filt}}$ & 0.14 & $1.11\times10^{-3}$ & 22.74 & $1.78\times10^{-1}$ \\
& & & & & & $i_{i}^{\mathrm{filt}}$ & 1.39 & $1.04\times10^{-2}$ & 12.48 & $9.30\times10^{-2}$ \\
\midrule
\multirow{4}{*}{\makecell[l]{(D2) $\lambda_{\mathrm{sim}}{=}0$}} &
\multirow{4}{*}{1H} & \multirow{4}{*}{4} & \multirow{4}{*}{$\checkmark$} &
\multirow{4}{*}{0} & \multirow{4}{*}{1.0} &
$i^{\mathrm{cv}}_d$ & 65.65 & $5.14\times10^{-1}$ & 0.04 & $2.98\times10^{-4}$ \\
& & & & & & $i^{\mathrm{cv}}_q$ & 18.14 & $1.35\times10^{-1}$ & 1.41 & $1.05\times10^{-2}$ \\
& & & & & & $i_{r}^{\mathrm{filt}}$ & 121.70 & $9.53\times10^{-1}$ & 0.04 & $3.13\times10^{-4}$ \\
& & & & & & $i_{i}^{\mathrm{filt}}$ & 24.95 & $1.86\times10^{-1}$ & 1.41 & $1.05\times10^{-2}$ \\
\bottomrule
\end{tabular}
\end{table*}

\subsection{Robustness Under Distribution Shift and Measurement Corruption}\label{subsec:gen}

Table~\ref{tab:gen} compares the $K=4$ and $K=1$ configurations under operating-point shifts that partially extend beyond the training range. We define the command deviation as $\Delta p_m=p_m-0.5$. The two shifted commands are $0.5+1.2\Delta p_m$ and $0.5+1.5\Delta p_m$, mapping the training interval $[0.4,0.8]$ to $[0.38,0.86]$ and $[0.35,0.95]$, respectively. Both shifted intervals extend beyond the training range at their lower and upper ends. Under clean conditions, $K=4$ has lower errors for more outputs, although neither configuration dominates across all outputs. At $\Delta p_m\!\times\!1.2$, the advantage is task- and channel-dependent. For $i_{r}^{\mathrm{filt}}$, $K=4$ gives a forecasting G-NRMSE of $0.24\%$, compared with $0.42\%$ for $K=1$, whereas $K=1$ gives a slightly lower simulation G-NRMSE for $i_d^{\mathrm{cv}}$. At $\Delta p_m\!\times\!1.5$, $K=4$ has lower errors across all outputs in both tasks. Thus, in this generalization experiment, the $K=4$ configuration has a broader advantage under the larger shift.

Sensor dropout is more damaging than additive measurement noise. With $p=0.2$ random dropout, forecasting $i^{\mathrm{cv}}_d$ reaches $4.10\%$, whereas doubling the measurement-noise scale gives $0.23\%$. Additive noise perturbs measurement magnitudes, but dropout removes parts of the sequential context. The stronger degradation of the $d/r$ outputs is consistent with Table~\ref{tab:accuracy}: the near-zero $q/i$ outputs have less G-NRMSE headroom relative to the zero-delta baseline.

\begin{table}[t]
\renewcommand{\arraystretch}{1.05}
\caption{Robustness under distribution shifts and measurement corruption. Entries are RMSE (pu) / G-NRMSE (\%). Dashes (---) mark settings not evaluated; the sensor-dropout and measurement-noise tests are forecasting-only.}
\label{tab:gen}
\centering
\scriptsize
\setlength{\tabcolsep}{3pt}
\begin{tabular}{l l l c c}
\toprule
\textbf{Category} & \textbf{Condition} & \textbf{Output}
& \textbf{Simulation} & \textbf{Forecasting} \\
\midrule
\multirow{4}{*}{Baseline} & \multirow{4}{*}{Clean} &
$i^{\mathrm{cv}}_d$ & $1.06{\times}10^{-3}/0.14$ & $7.93{\times}10^{-4}/0.10$ \\
& & $i^{\mathrm{cv}}_q$ & $1.04{\times}10^{-2}/1.39$ & $1.05{\times}10^{-2}/1.41$ \\
& & $i_{r}^{\mathrm{filt}}$ & $1.64{\times}10^{-3}/0.21$ & $9.70{\times}10^{-4}/0.12$ \\
& & $i_{i}^{\mathrm{filt}}$ & $1.04{\times}10^{-2}/1.39$ & $1.05{\times}10^{-2}/1.40$ \\
\midrule
\multirow{16}{*}{\makecell[l]{Operating-point\\shift}} &
\multirow{4}{1.4cm}{$\Delta p_m\!\times\!1.2$ \\ ($K{=}4$)} &
$i^{\mathrm{cv}}_d$ & $1.68{\times}10^{-3}/0.22$ & $2.67{\times}10^{-3}/0.34$ \\
& & $i^{\mathrm{cv}}_q$ & $2.69{\times}10^{-2}/3.62$ & $2.72{\times}10^{-2}/3.65$ \\
& & $i_{r}^{\mathrm{filt}}$ & $2.37{\times}10^{-3}/0.30$ & $1.85{\times}10^{-3}/0.24$ \\
& & $i_{i}^{\mathrm{filt}}$ & $2.69{\times}10^{-2}/3.61$ & $2.72{\times}10^{-2}/3.65$ \\
\cmidrule(lr){2-5}
& \multirow{4}{1.4cm}{$\Delta p_m\!\times\!1.5$ \\ ($K{=}4$)} &
$i^{\mathrm{cv}}_d$ & $2.17{\times}10^{-2}/2.77$ & $4.13{\times}10^{-3}/0.53$ \\
& & $i^{\mathrm{cv}}_q$ & $3.93{\times}10^{-2}/5.28$ & $3.14{\times}10^{-2}/4.22$ \\
& & $i_{r}^{\mathrm{filt}}$ & $2.19{\times}10^{-2}/2.80$ & $3.01{\times}10^{-3}/0.38$ \\
& & $i_{i}^{\mathrm{filt}}$ & $3.95{\times}10^{-2}/5.30$ & $3.14{\times}10^{-2}/4.21$ \\
\cmidrule(lr){2-5}
& \multirow{4}{1.4cm}{$\Delta p_m\!\times\!1.2$ \\ ($K{=}1$)} &
$i^{\mathrm{cv}}_d$ & $1.40{\times}10^{-3}/0.18$ & $3.01{\times}10^{-3}/0.38$ \\
& & $i^{\mathrm{cv}}_q$ & $2.70{\times}10^{-2}/3.63$ & $2.73{\times}10^{-2}/3.66$ \\
& & $i_{r}^{\mathrm{filt}}$ & $1.73{\times}10^{-3}/0.22$ & $3.29{\times}10^{-3}/0.42$ \\
& & $i_{i}^{\mathrm{filt}}$ & $2.70{\times}10^{-2}/3.62$ & $2.73{\times}10^{-2}/3.66$ \\
\cmidrule(lr){2-5}
& \multirow{4}{1.4cm}{$\Delta p_m\!\times\!1.5$ \\ ($K{=}1$)} &
$i^{\mathrm{cv}}_d$ & $2.60{\times}10^{-2}/3.31$ & $5.26{\times}10^{-3}/0.67$ \\
& & $i^{\mathrm{cv}}_q$ & $4.33{\times}10^{-2}/5.81$ & $3.16{\times}10^{-2}/4.24$ \\
& & $i_{r}^{\mathrm{filt}}$ & $2.75{\times}10^{-2}/3.51$ & $5.25{\times}10^{-3}/0.67$ \\
& & $i_{i}^{\mathrm{filt}}$ & $4.31{\times}10^{-2}/5.78$ & $3.16{\times}10^{-2}/4.24$ \\
\midrule
\multirow{4}{*}{\makecell[l]{Sensor\\Dropout\\($p{=}0.2$)}} &
\multirow{4}{1.4cm}{Forecasting \\ only} &
$i^{\mathrm{cv}}_d$ & --- & $3.21{\times}10^{-2}/4.10$ \\
& & $i^{\mathrm{cv}}_q$ & --- & $1.35{\times}10^{-2}/1.81$ \\
& & $i_{r}^{\mathrm{filt}}$ & --- & $3.20{\times}10^{-2}/4.09$ \\
& & $i_{i}^{\mathrm{filt}}$ & --- & $1.29{\times}10^{-2}/1.73$ \\
\midrule
\multirow{8}{*}{\makecell[l]{Measurement\\Noise}} &
\multirow{4}{*}{$\times\!1.5$ scale} &
$i^{\mathrm{cv}}_d$ & --- & $1.49{\times}10^{-3}/0.19$ \\
& & $i^{\mathrm{cv}}_q$ & --- & $1.05{\times}10^{-2}/1.41$ \\
& & $i_{r}^{\mathrm{filt}}$ & --- & $1.56{\times}10^{-3}/0.20$ \\
& & $i_{i}^{\mathrm{filt}}$ & --- & $1.05{\times}10^{-2}/1.41$ \\
\cmidrule(lr){2-5}
& \multirow{4}{*}{$\times\!2.0$ scale} &
$i^{\mathrm{cv}}_d$ & --- & $1.83{\times}10^{-3}/0.23$ \\
& & $i^{\mathrm{cv}}_q$ & --- & $1.05{\times}10^{-2}/1.41$ \\
& & $i_{r}^{\mathrm{filt}}$ & --- & $1.88{\times}10^{-3}/0.24$ \\
& & $i_{i}^{\mathrm{filt}}$ & --- & $1.05{\times}10^{-2}/1.41$ \\
\bottomrule
\end{tabular}
\end{table}

\subsection{Cross-Topology Forecasting and Adaptation: 9-Bus System}
\label{subsec:ninebus}

Next, we evaluate cross-topology forecasting and target-domain adaptation on a 9-bus system. The network contains one dynamically modeled synchronous generator, a grid-following inverter, and a fixed infinite bus as the slack reference. The 9-bus system uses a 100-MVA base. For its inverter model, $r_f=0.016$ and $x_f=0.009$\,pu are specified directly on this base, not converted from the 2.75-MVA base of the offline benchmark. Trajectories are integrated with a fixed-step fourth-order Runge--Kutta method using a 1-ms internal step and recorded every 5\,ms.

Within the 9-bus domain, the in-distribution (ID) set and two out-of-distribution sets (OOD-1 and OOD-2) contain 200, 50, and 50 trajectories, with $p_m$ ranges of $[0.3,0.7]$, $[0.7,0.9]$, and $[0.9,1.1]$\,pu, respectively. Fine-tuning (FT) uses 160 ID trajectories; the remaining 40 form the validation subset reported as ID in Table~\ref{tab:ninebus}, while both OOD sets are reserved for evaluation. The zero-shot (ZS) source model and the fine-tuned model are evaluated on the same subsets. Table~\ref{tab:ninebus} reports G-NRMSE normalized by each target set's per-channel range; comparisons are therefore made within each set. Persistence repeats the latest observed output as the 5-ms-ahead forecast, $\hat{x}_{t+1}=x_t$. ZS has higher error than persistence in all 12 set--channel comparisons. FT matches or outperforms persistence in 11 of the 12 comparisons; the only exception is $i_{r}^{\mathrm{filt}}$ on OOD-2 ($2.11\%$ versus $2.08\%$). Across the evaluated sets and channels, FT lowers the ZS error in all 12 comparisons.

\subsection{Cross-System Adaptation to CHIL Measurements}\label{subsec:hil}

To assess cross-system adaptation, we evaluated Mamba--MoE using measurement data collected from a CHIL simulation of a three-phase grid-tied inverter \cite{Orsinger2021}. Its four channels, $[i_{d}^{\mathrm{cv}}, i_{q}^{\mathrm{cv}}, i_{d}^{\mathrm{filt}}, i_{q}^{\mathrm{filt}}]^\mathsf{T}$, form two measured-current pairs in the PLL-aligned $dq$ frame, whereas the offline outputs represent one branch current in two frames. The listed order defines the channel mapping. The logged PLL angle was used upstream to form the recorded $dq$ channels but was neither a surrogate input nor an output. The 1-kHz record contains DC-link-reference changes and was downsampled fivefold to 200\,Hz.

\begin{table}[t]
\renewcommand{\arraystretch}{0.95}
\caption{Five-ms-ahead forecasting on the 9-bus system (target-set G-NRMSE, \%).}
\label{tab:ninebus}
\centering
\scriptsize
\setlength{\tabcolsep}{2.5pt}
\begin{tabular}{@{}l l c c c c@{}}
\toprule
\textbf{Set} & \textbf{Method} & $i^{\mathrm{cv}}_d$ & $i^{\mathrm{cv}}_q$ & $i_{r}^{\mathrm{filt}}$ & $i_{i}^{\mathrm{filt}}$ \\
\midrule
\multirow{3}{*}{ID}
& Persistence & 2.23 & 3.01 & 2.16 & 3.15 \\
& ZS & 3.21 & 5.29 & 3.07 & 5.97 \\
& FT & 1.55 & 2.57 & 1.26 & 2.68 \\
\midrule
\multirow{3}{*}{OOD-1}
& Persistence & 1.18 & 1.72 & 1.16 & 1.52 \\
& ZS & 3.11 & 16.32 & 4.84 & 15.61 \\
& FT & 0.96 & 1.72 & 0.90 & 1.47 \\
\midrule
\multirow{3}{*}{OOD-2}
& Persistence & 2.02 & 2.86 & 2.08 & 2.41 \\
& ZS & 2.22 & 15.29 & 4.05 & 13.55 \\
& FT & 1.82 & 2.77 & 2.11 & 2.32 \\
\bottomrule
\end{tabular}
\end{table}

\begin{figure}[t]
\centering
\includegraphics[width=.9\columnwidth]{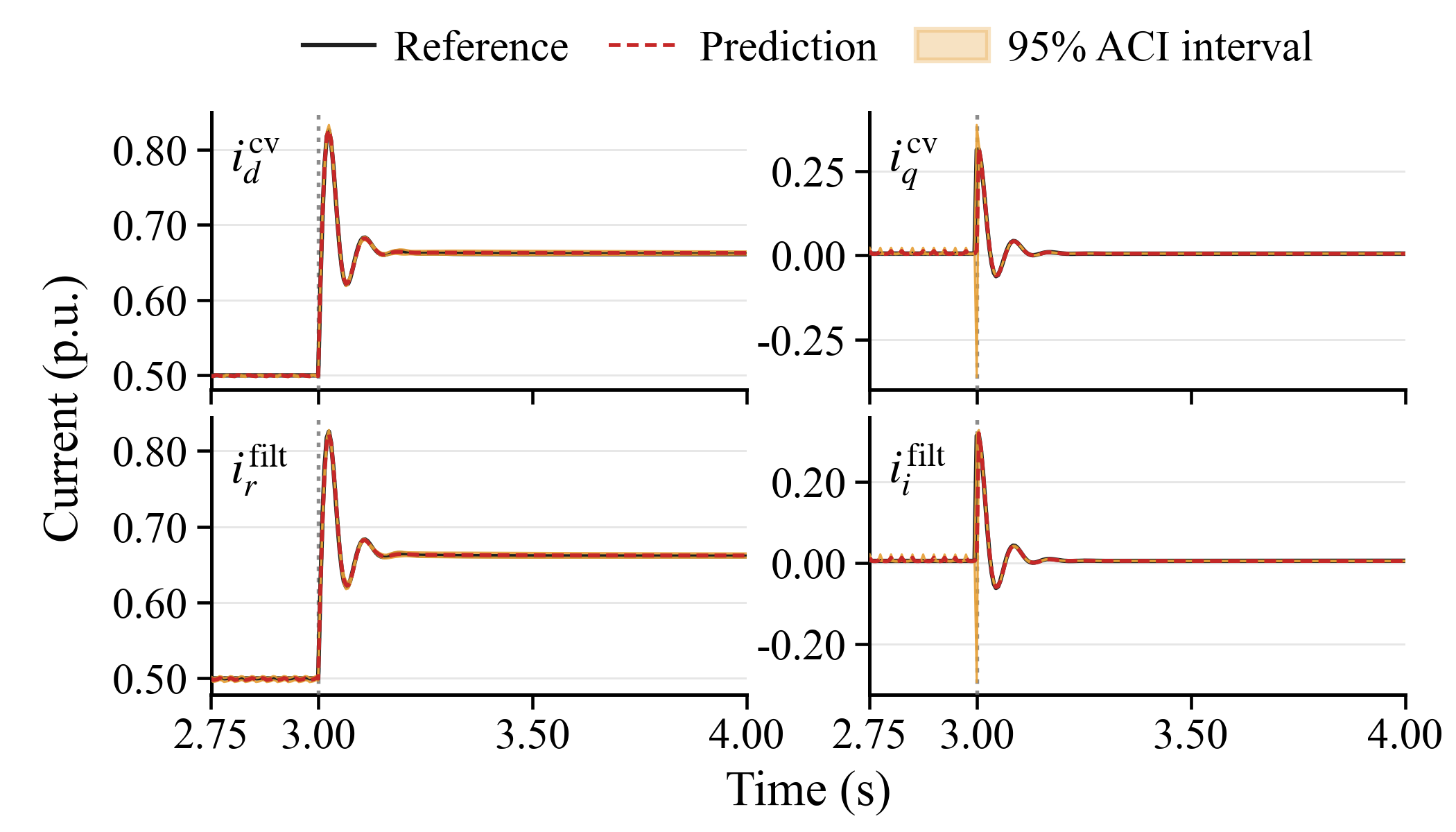}
\vspace{-.4cm}
\caption{Simulation predictions with ACI intervals under a step disturbance.}
\label{fig:sim_traj}
\end{figure}

\begin{figure}[t]
\centering
\includegraphics[width=.9\columnwidth]{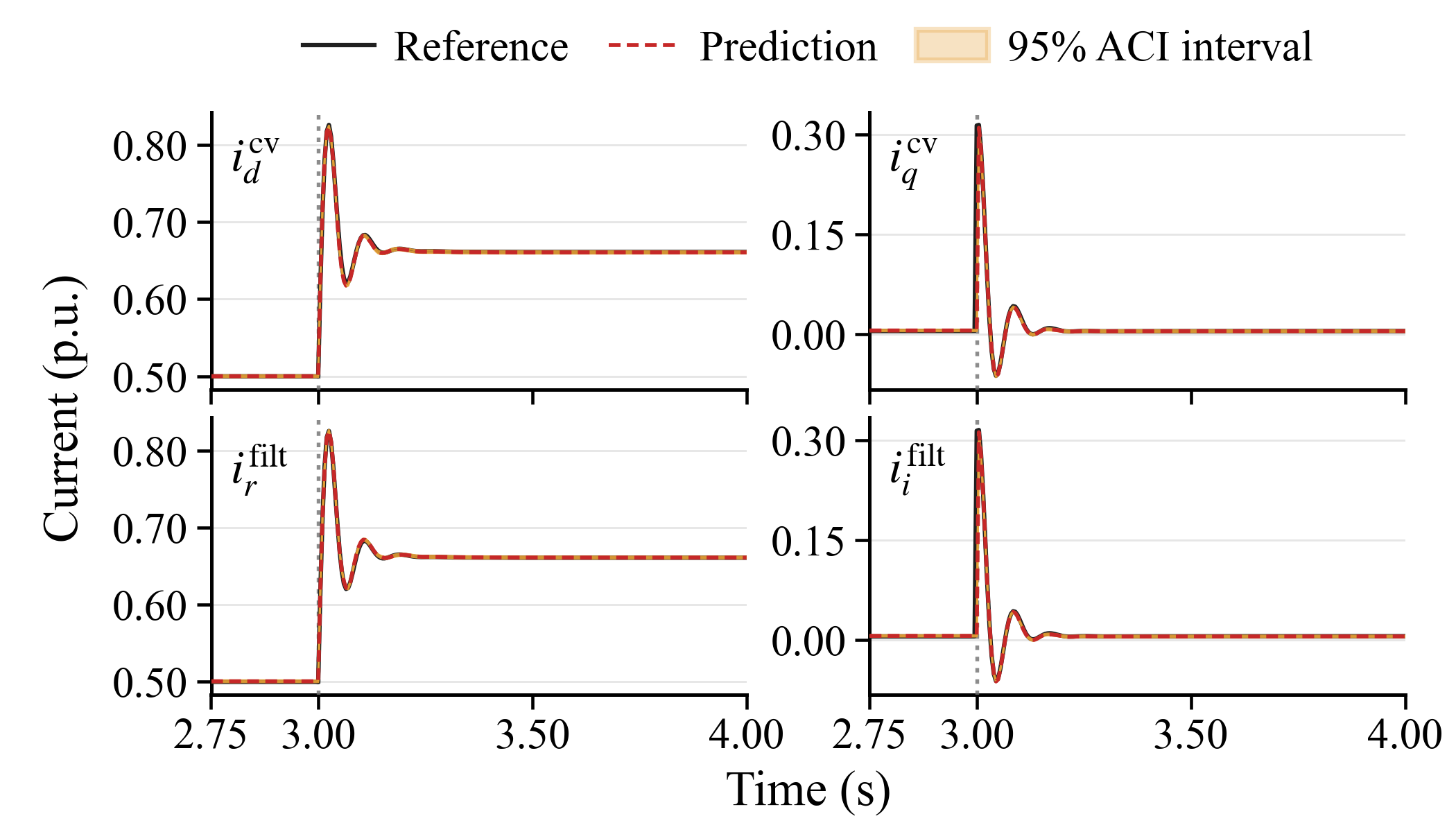}
\vspace{-.4cm}
\caption{Forecasting predictions with ACI intervals.}
\label{fig:forec_uq}
\end{figure}

\begin{figure}[t]
\centering
\includegraphics[width=.9\columnwidth]{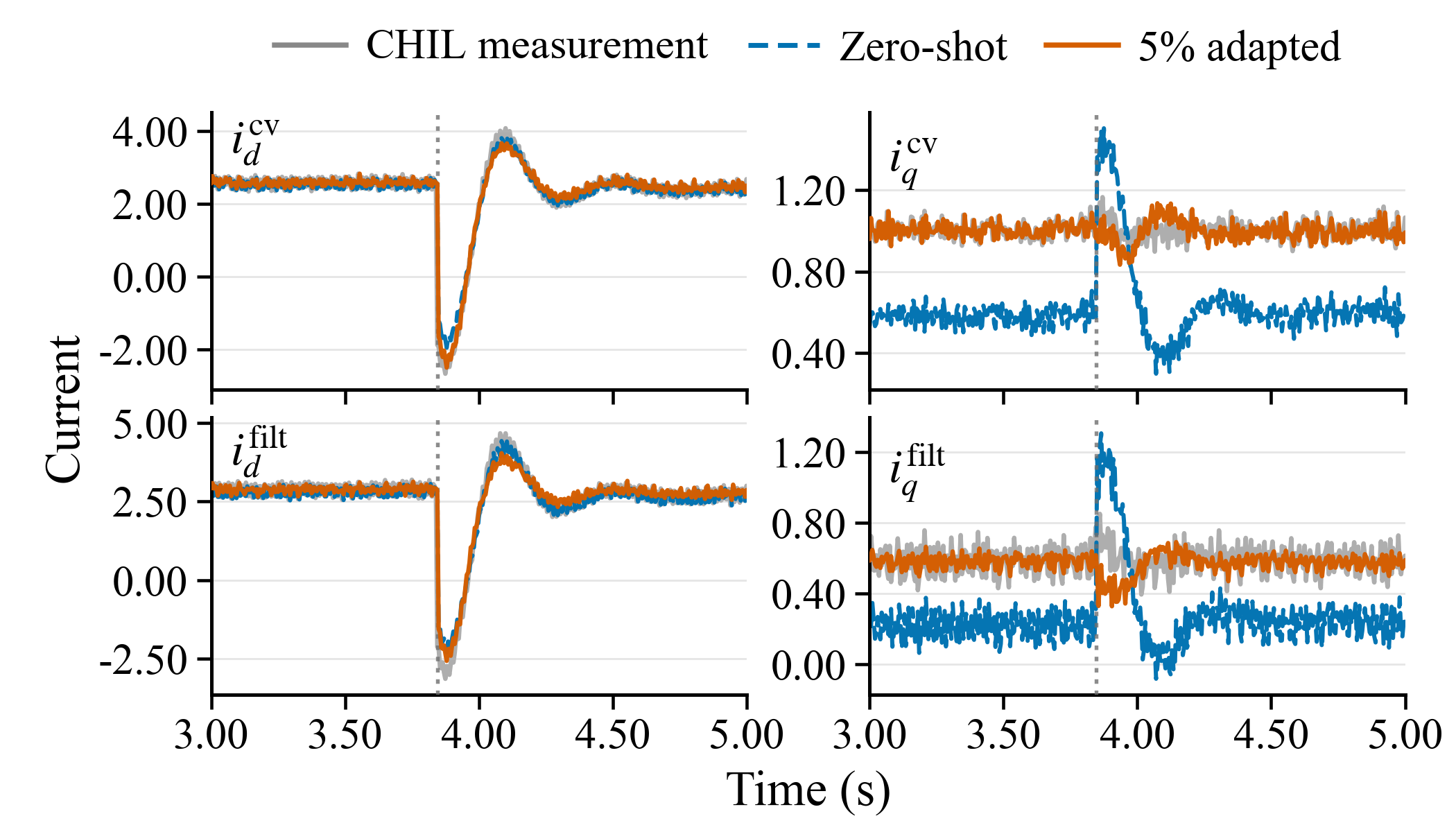}
\vspace{-.4cm}
\caption{CHIL current forecasts before and after 10-s (5.1\%)
shared-output-head adaptation; the vertical reference line marks a
DC-link-reference change.}
\label{fig:hil_forec}
\end{figure}

We compared the frozen source model with one whose shared output head was adapted using the first 10\,s (5.1\%) and only the first-horizon loss. Both were evaluated 5\,ms ahead on the fixed test block beginning at 42\,s. The CHIL values were not converted to the offline benchmark's pu base. After normalization by each channel's test-block range, the mean normalized RMSE decreased from 13.09\% to 3.07\%. Fig.~\ref{fig:hil_forec} compares the measurements with the frozen and adapted forecasts.

\subsection{Interval Reliability: Static CP and ACI}

Table~\ref{tab:aci} compares static CP and ACI. Simulation residuals are more nonstationary than forecasting residuals because rollout errors evolve across time, whereas closest-window merging produces more homogeneous forecasting residuals. For simulation, static CP achieves $91.70\%$ MCOV, below the $95\%$ target. ACI raises MCOV to $94.47\%$ and reduces mean MPIW from $0.005648$ to $0.004823$. For forecasting, static CP already reaches $94.55\%$ MCOV with mean MPIW $0.002072$. ACI shifts coverage to $96.04\%$ with a small width increase to $0.002255$. The practical benefit of ACI is therefore most evident for simulation. The remaining simulation coverage gap is about 0.5 percentage points and occurs near disturbance onset. At that point, the operating point changes abruptly and the residual grows before the online scale factor $\kappa_u(j)$ accumulates enough feedback. Fig.~\ref{fig:sim_traj} shows simulation predictions and ACI 95\% intervals. The intervals widen near $t=3$\,s, where the disturbance causes the largest transient residuals. Fig.~\ref{fig:forec_uq} shows the forecasting counterpart. After closest-window merging, the forecasting intervals are narrower and more uniform across target times than the simulation intervals.

\begin{table}[t]
\renewcommand{\arraystretch}{1.1}
\caption{ACI interval reliability.}
\label{tab:aci}
\centering
\scriptsize
\setlength{\tabcolsep}{3pt}
\begin{tabular}{llcc}
\toprule
\textbf{Method} & \textbf{Mode}
& \makecell{\textbf{Mean MCOV}\\\textbf{(\%)}} & \makecell{\textbf{Mean MPIW}\\\textbf{(pu)}} \\
\midrule
CP (symmetric) & Simulation & 91.70 & 0.005648 \\
ACI (symmetric) & Simulation & 94.47 & 0.004823 \\
\midrule
CP (symmetric) & Forecasting & 94.55 & 0.002072 \\
ACI (symmetric) & Forecasting & 96.04 & 0.002255 \\
\bottomrule
\end{tabular}
\end{table}

\section{Conclusion}
\label{sec:conclusion}

This paper studies a unified surrogate for autoregressive simulation and measurement-window forecasting of sampled closed-loop current outputs. The tasks share the same current-output channels but differ in prediction form: autoregressive rollout versus single-pass multi-step forecasting from a measurement window. Mamba--MoE learns a single temporal representation for shared transient dynamics, while task conditioning and expert routing adapt task-dependent residuals. Task-matched objectives address error propagation, and ACI calibration improves interval reliability.

On the inverter benchmark, Mamba--MoE keeps errors for both tasks close to those of a Mamba specialist pair while reducing parameters by about 13\%. ACI intervals achieve about 94--96\% mean marginal coverage, close to the 95\% target. At the larger operating-point shift evaluated in this study, expert routing yields lower errors across all outputs in both tasks. In the cross-system CHIL study, shared-output-head adaptation with limited measured data reduces the mean held-out forecasting error across the four measured-current channels. Future work will extend the framework to multi-inverter networks, stability-constrained long-horizon rollout, and physics-guided constraint handling for stiff differential--algebraic systems~\cite{le2026physicsguided}.


\end{document}